%% file: main.tex
\documentclass[pdflatex,sn-mathphys,referee,Numbered]{sn-jnl}
\usepackage{nature}
\graphicspath{{figures/NMI}}
\usepackage{anyfontsize}
\usepackage{siunitx}
\begin{document}
\input{config}

\title[test]{\centering\Title\footnote{A video demonstration of the system is available on \href{\suppUrlnmi}{Vimeo} (up to 4k, with English subtitle). The complete URL is \url{\suppUrlnmi}.}}

\author[1,2]{\fnm{Zihang} \sur{Zhao}}\email{zhaozihang@stu.pku.edu.cn}
\equalcont{These authors contributed equally to this work.}

\author[2]{\fnm{Wanlin} \sur{Li}}\email{liwanlin@bigai.ai}
\equalcont{These authors contributed equally to this work.}

\author[1,2]{\fnm{Yuyang} \sur{Li}}\email{y.li@stu.pku.edu.cn}
\equalcont{These authors contributed equally to this work.}

\author[2]{\fnm{Tengyu} \sur{Liu}}\email{liutengyu@bigai.ai}
\equalcont{These authors contributed equally to this work.}

\author[2]{\fnm{Boren} \sur{Li}}\email{liboren@bigai.ai}

\author[2]{\fnm{Meng} \sur{Wang}}\email{wangmeng@bigai.ai}

\author[1]{\fnm{Kai} \sur{Du}}\email{kai.du@pku.edu.cn}

\author*[2]{\fnm{Hangxin} \sur{Liu}}\email{liuhx@bigai.ai}

\author*[1,3]{\fnm{Yixin} \sur{Zhu}}\email{yixin.zhu@pku.edu.cn}

\author[4]{\fnm{Qining} \sur{Wang}}\email{qiningwang@pku.edu.cn}

\author*[5]{\fnm{Kaspar} \sur{Althoefer}}\email{k.althoefer@qmul.ac.uk}

\author[1,2]{\fnm{Song-Chun} \sur{Zhu}}\email{s.c.zhu@pku.edu.cn}

\affil[1]{\orgdiv{Institute for Artificial Intelligence}, \orgname{Peking University}, \orgaddress{\street{5 Yiheyuan Road}, \city{Haidian}, \postcode{100871}, \state{Beijing}, \country{China}}}

\affil[2]{\orgname{Beijing Institute for General Artificial Intelligence}, \orgaddress{\street{2 Yiheyuan Road}, \city{Haidian}, \postcode{100080}, \state{Beijing}, \country{China}}}

\affil[3]{\orgname{PKU-Wuhan Institute for Artificial Intelligence}, \orgaddress{\street{770 Gaoxin Road}, \city{Wuhan}, \postcode{430075}, \state{Hubei}, \country{China}}}

\affil[4]{\orgdiv{College of Engineering}, \orgname{Peking University}, \orgaddress{\street{5 Yiheyuan Road}, \city{Haidian}, \postcode{100871}, \state{Beijing}, \country{China}}}

\affil[5]{\orgdiv{School of Engineering and Materials Science}, \orgname{Queen Mary University of London}, \orgaddress{\street{Mile End Road}, \city{London}, \postcode{E1 4NS}, \country{UK}}}

\abstract{
Developing robotic hands that adapt to real-world dynamics remains a fundamental challenge in robotics and machine intelligence.
Despite significant advances in replicating human hand kinematics and control algorithms, robotic systems still struggle to match human capabilities in dynamic environments, primarily due to inadequate tactile feedback.
To bridge this gap, we present the \HandName{}, a biomimetic hand featuring high-resolution tactile sensing (\(0.1~\textrm{mm}\) spatial resolution) across \(70\%\) of its surface area.
Through optimized hand design, we overcome traditional challenges in integrating high-resolution tactile sensors while preserving the full range of motion. The hand, powered by our generative algorithm that synthesizes human-like hand configurations, demonstrates robust grasping capabilities in dynamic real-world conditions.
Extensive evaluation across \(600\) real-world trials demonstrates that this tactile-embodied system significantly outperforms non-tactile-informed alternatives in complex manipulation tasks (\(\text{p}<0.0001\)).
These results provide empirical evidence for the critical role of rich tactile embodiment in developing advanced robotic intelligence, offering promising perspectives on the relationship between physical sensing capabilities and intelligent behavior.}

\maketitle

\clearpage

Precise sensory-motor control in real-world scenarios is fundamental to machine intelligence and embodied \acf{ai}~\cite{brooks1991intelligence, segal2019more}. A hallmark challenge in this field is the control of dexterous robotic hands~\cite{billard2024good}. Despite advances in mechatronic systems and sophisticated finger designs that enable enhanced dexterity~\cite{ma2011dexterity}, the limited availability of rich sensory feedback fundamentally restricts their ability to adapt during dynamic interactions~\cite{billard2019trends,lepora2024future}. Understanding and addressing this sensory limitation is crucial for deploying robotic hands in real-world scenarios that demand nuanced control and rapid adaptation.

The robotics community has long recognized this challenge, approaching it through increasingly sophisticated hardware and control strategies. On the hardware front, researchers have developed intricate mechanical designs that closely mimic human hand kinematics~\cite{jacobsen1984utah,deimel2016novel,hughes2018anthropomorphic,de20223d,shadowrobot}, primarily relying on proprioceptive sensing for joint-level feedback. These hardware advances, often combined with visual perception, have enabled various control paradigms: from planning-based methods that execute precise finger gaiting~\cite{morgan2022complex,li2024grasp}, to learning-based approaches that develop control policies through training~\cite{andrychowicz2020learning,qin2022dexmv,chen2023visual}, and recently, to \acp{llm} that provide high-level task reasoning~\cite{ma2024eureka}. However, a fundamental limitation persists: without the direct sensation of local contacts---crucial information for both modeling and control---these systems fail to handle unexpected physical interactions~\cite{billard2019trends}.

The solution may lie in understanding human hand control, which achieves remarkable precise control through a sophisticated tactile perception system. This biological system comprises two key elements: a dense array of tactile sensors embedded throughout the skin~\cite{westling1984factors,johansson2009coding}, and specialized neural processing in the primary somatosensory cortex that rapidly interprets and integrates this massive sensory input~\cite{penfield1937somatic,kaas1979multiple,johansson2009coding}. This combination enables humans to instantly detect and respond to subtle contact changes during manipulation, a capability that current robotic systems have yet to replicate.

Drawing direct inspiration from this biological architecture, we present \HandName{} (Full-hand TACtile-embedded Biomimetic Hand), a system that bridges the sensory gap in robotic manipulation. The core innovation lies in its comprehensive tactile sensing capability, featuring high-resolution (\(0.1~\textrm{mm}\) spatial resolution) coverage across \(70~\%\) of the hand surface. This is achieved through the effective integration of 17 vision-based tactile sensors in six optimized configurations, where sensor covers serve dual purposes as both sensing elements and structural components. The hand maintains full human-like dexterity, demonstrated by its high Kapandji score~\cite{kapandji1986clinical} and ability to perform all 33 human grasp types~\cite{feix2015grasp}. Complementing this hardware, we developed a generative algorithm that produces human-like hand configurations, creating a rich knowledge base for object interaction. The integration enables closed-loop tactile-informed control that processes high-dimensional contact data for precise, adaptive manipulation.

To rigorously validate \HandName{'s} capabilities, we focused on multi-object grasping---a task that epitomizes the challenges of dexterous manipulation~\cite{billard2019trends,yao2023exploiting}. While single-object manipulation has been successfully addressed by \(1\)-\ac{dof} parallel grippers~\cite{she2021cable,lloyd2024pose,zhao2024tac}, simultaneous manipulation of multiple objects presents two distinct challenges: it requires both precise contact detection across the entire hand and strategic motion adjustments to prevent object collisions. Through comprehensive tactile sensing, \HandName{} directly addresses these challenges. Extensive evaluation across \(600\) real-world trials demonstrates significant performance improvements over non-tactile alternatives (\(\text{p}<0.0001\)), particularly in scenarios involving real-world execution noise and dynamic object interactions.

Our work advances the field through two primary contributions: a practical demonstration that full-hand tactile sensing can be achieved without compromising hand motion capabilities, and comprehensive empirical validation of its benefits. By solving the technical challenges that previously restricted tactile sensing to simple grippers, this research enables unprecedented investigations into sophisticated tactile-embodied intelligence~\cite{lepora2024future}. More broadly, our results provide concrete evidence for the critical role of rich sensory feedback in intelligent behavior, suggesting promising directions for developing embodied \ac{ai} systems beyond purely computational approaches~\cite{Turing1950computing,mitchell2024debates}.

\clearpage
\section*{Results}

\subsection*{\HandName{} hardware}

\begin{figure}[t!]
    \centering
    \includegraphics[width=\linewidth]{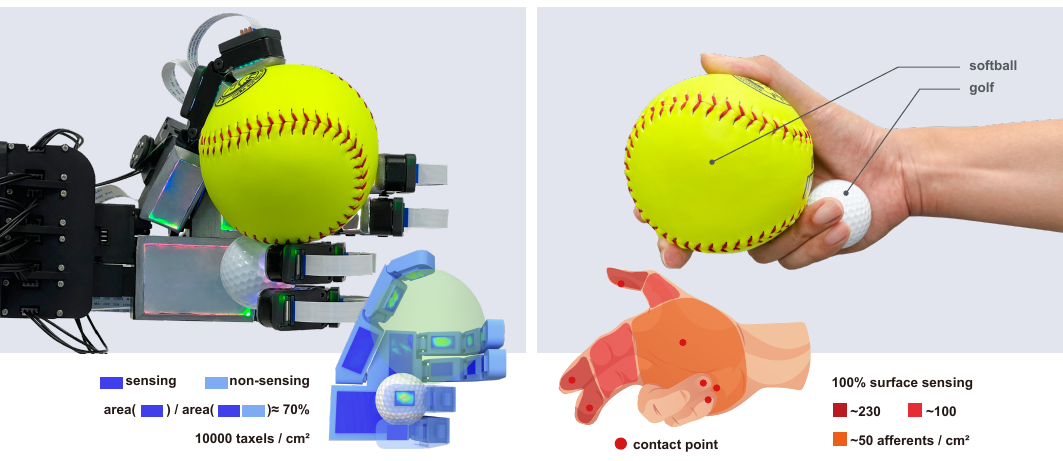}
    \caption{\textbf{\textbar~ Overview of the \HandName{}.} \HandName{} is a dexterous robotic hand featuring a high-density tactile sensing array that matches human capabilities, as benchmarked against physiological data from Vallbo~\etal~\cite{vallbo1984properties}. Detailed illustrations of the sensor construction and assembly process are provided in \cref{fig:results_hardware} and \ref{fig:extended_mechatronics_design}. Similar to its biological counterpart, it leverages sophisticated tactile feedback to accomplish complex manipulation tasks, such as precise in-hand object pose arrangement, enabling simultaneous and stable grasping of multiple items, a capability highlighted as challenging but crucial by Billard~\etal~\cite{billard2019trends}}.
    \label{fig:results_overview}
\end{figure}

\begin{figure}[t!]
    \centering
    \includegraphics[width=0.96\linewidth]{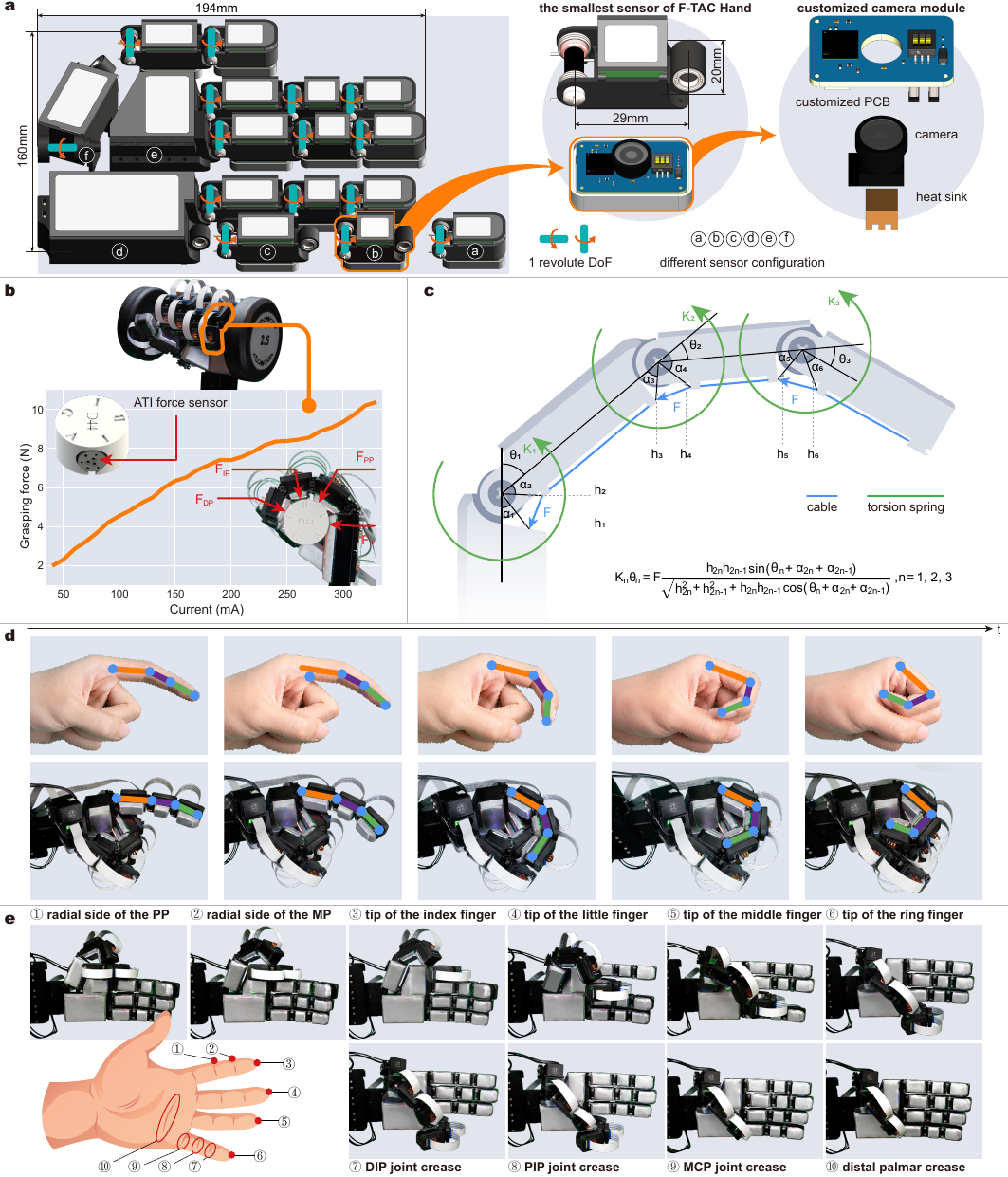}
    \caption{\textbf{Hardware of the \HandName{}.} \textbf{a}, The seamless integration of $17$ vision-based sensors in $6$ configurations, maintaining $15$ \acp{dof}---three per finger---and adult hand dimensions. Each sensor includes a streamlined camera module for efficient tactile data acquisition in confined space. \textbf{b}, \HandName{} demonstrates its strength by holding a $2.5~\mathrm{kg}$ dumbbell; each phalanx contributes to a total grasping force of $10.3~\mathrm{N}$. \textbf{c}, Schematic representation of a finger, with $K_n$, $\theta_n$, and $F$ denoting joint stiffness, rotation angle, and cable force, respectively. Offsets in rotation due to cable and joint alignment are also shown. \textbf{d}, Top-down view comparison of \HandName{} and human finger flexion. \textbf{e}, Despite the numerous sensors, \HandName{} retains its mobility, as evidenced by a successful Kapandji test~\cite{kapandji1986clinical}, where the thumb fingertip sequentially touches specific points on the hand as numbered in the figure.}
    \label{fig:results_hardware}
\end{figure}

\HandName{} advances the state of dexterous robotic hands through its comprehensive tactile sensing capabilities while maintaining a full range of motion. The hand achieves human-like tactile coverage, with sensing elements extending across $70~\%$ of the palmar surface at a density of \(\num{10000}\) taxels---the pixels in camera CMOS---\(/\mathrm{cm}^2\) (\cref{fig:results_overview}), significantly surpassing current commercial solutions like the Shadow hand, which provides only five-point feedback over less than $20~\%$ of its surface~\cite{shadowrobot} (comparison with other tactile arrays is available at Supplementary Information \cref{sec:supp:sensor_characteristics}). This extensive coverage is achieved through an array of vision-based tactile sensors in multiple configurations (see exploded view in \ref{fig:extended_mechatronics_design}a and physical dimensions in Supplementary Information \cref{sec:supp:sensor_characteristics}), featuring specially designed covers that align with the hand's phalanges and palm to minimize mechanical redundancy while replicating the natural kinematic structure of the human hand (\cref{fig:results_hardware}a). Each of the five fingers incorporates three \acp{dof}, contributing to the hand's total 15 \acp{dof} configuration that enables human-like dexterity. A specialized electronic module enables large-scale sensor reading acquisition while minimizing space, weight, and cabling requirements (\ref{fig:extended_mechatronics_design}b). The hand's dimensions mirror those of an adult human hand, measuring $194~\mathrm{mm}$ from wrist to middle fingertip (\cref{fig:results_hardware}a), and its modular design architecture allows for easy adaptation to different physical dimensions while maintaining functionality.

\begin{figure}[t!]
    \centering
    \includegraphics[width=\linewidth]{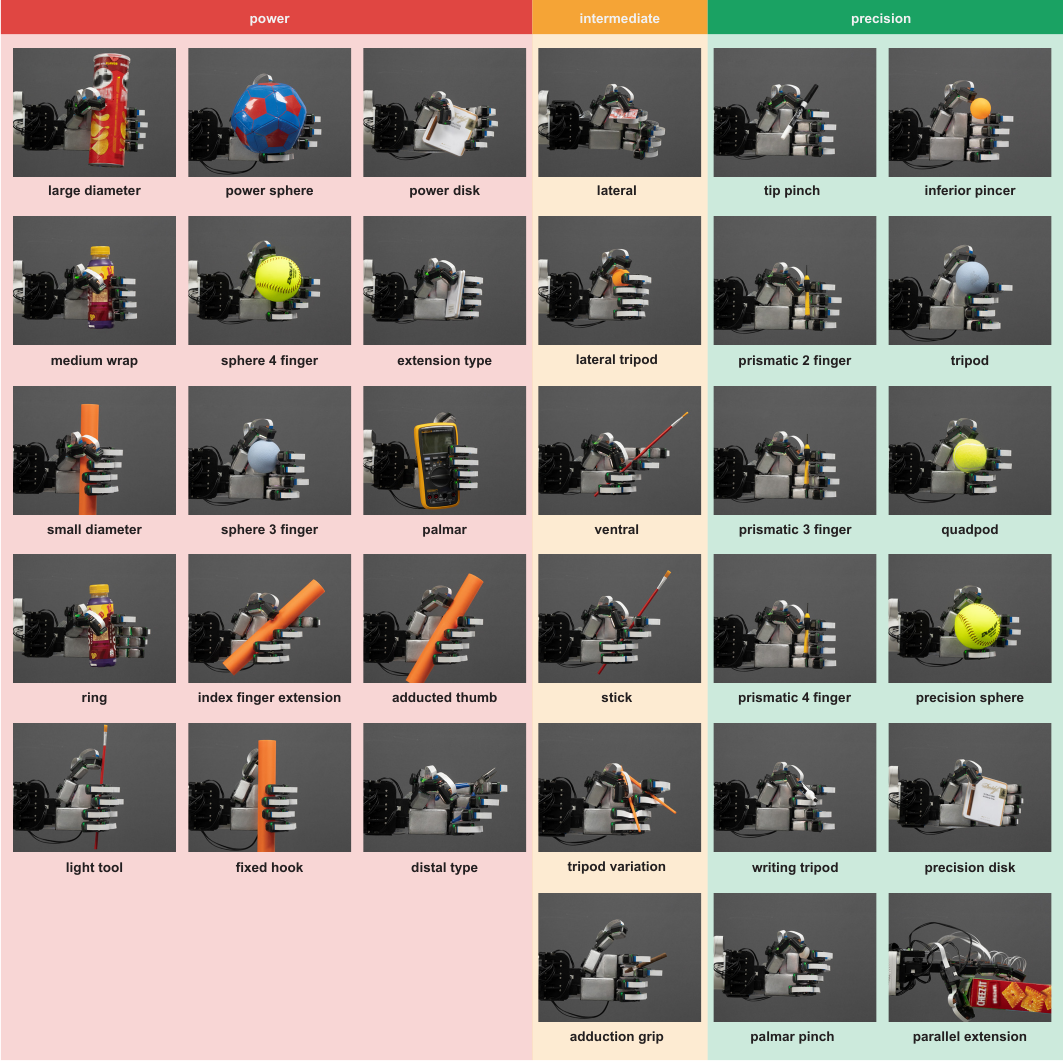}
    \caption{\textbf{\textbar~ Workspace of the \HandName{}.} Empowered by its smart design, the workspace of the \HandName{} enables it to perform all $33$ human grasping types, as documented by Feix~\etal~\cite{feix2015grasp}.}
    \label{fig:results_workspace}
\end{figure}

Building upon its extensive tactile sensing coverage, \HandName{} also achieves comprehensive motion capabilities that match state-of-the-art dexterous hands~\cite{hughes2018anthropomorphic,de20223d,shadowrobot}. The hand implements full mobility using just five slim cables (\ref{fig:extended_mechatronics_design}c) with substantial payload capacity (\cref{fig:results_hardware}b). Each cable controls the flexion and extension of a finger (\ref{fig:extended_mechatronics_design}d--e), working in concert with stiffness-tuned springs at each joint (\cref{fig:results_hardware}c) to replicate the coordinated yet semi-independent movements characteristic of human hands~\cite{sancho20013} (\cref{fig:results_hardware}d). An additional degree of actuation enables thumb opposition, expanding the hand's motion versatility (\ref{fig:extended_mechatronics_design}e). Detailed fabrication procedures are provided in `Tactile sensor fabrication' and `\HandName{} fabrication' in \hyperref[sec:methods]{Methods}. The hand's dexterity is demonstrated through two evaluations: the Kapandji test~\cite{kapandji1986clinical}, completing all 10 designated thumb-to-hand contact points shown in \cref{fig:results_hardware}e, and the successful execution of all \(33\) human grasp types (\cref{fig:results_workspace}).

\begin{figure}[t!]
    \centering
    \includegraphics[width=\linewidth]{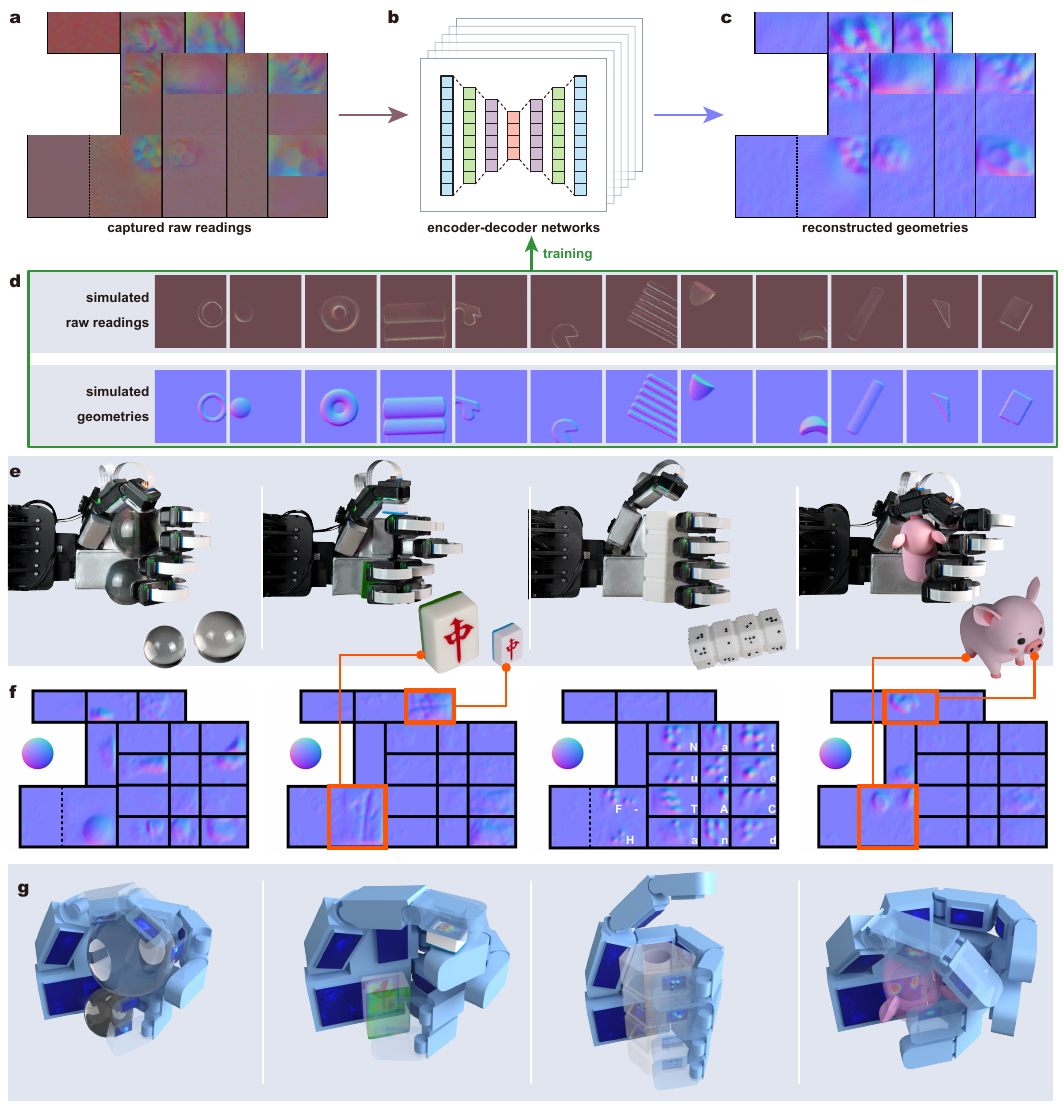}
    \caption{\textbf{\textbar~ Comprehensive tactile sensing capabilities of the \HandName{}.} \textbf{a}, Raw tactile sensor readings from the configuration shown in \cref{fig:results_overview}\textbf{a} are processed by neural networks (\textbf{b}) to reconstruct contact site geometries (\textbf{c}), visualized as normal maps. The neural networks, trained on simulated data (\textbf{d}) generated by a physics-based image formation model, enable efficient and precise mapping of extensive raw data to geometric information at the contact interface. When grasping an object (\textbf{e}), \HandName{} captures detailed contact information through its advanced tactile sensing capabilities (\textbf{f}). \textbf{g}, This rich tactile feedback enables \HandName{} to accurately perceive and interpret object characteristics, as demonstrated by its precise estimation of in-hand object poses.}
    \label{fig:results_tactile}
\end{figure}

The hand's tactile sensing system utilizes the photometric stereo principle~\cite{woodham1980photometric,yuan2017gelsight}, converting light intensity variations into surface gradient information (\cref{fig:results_tactile}a). Contact surface geometry is reconstructed through a two-stage process. First, an array of encoder-decoder neural networks (\cref{fig:results_tactile}b) maps physics-based relationships between surface gradients and intensity variations for each sensor. Next, a Poisson solver generates high-fidelity surface geometries, visualized as normal maps (\cref{fig:results_tactile}c). The detailed sensor characteristics are available in the Supplementary Information \cref{sec:supp:sensor_characteristics}.

The unprecedented scale of \HandName{'s} tactile sensing system required developing efficient calibration solutions. We addressed this through a physics-based image formation model (detailed in Supplementary Information \cref{sec:supp:image_formation} and \cref{sec:supp:network}) that generates synthetic readings of elastomer deformations during contacts (\cref{fig:results_tactile}d). This approach enables efficient neural network training (\cref{fig:results_tactile}b) and accurate sensor calibration.

The integration of fine-grained tactile sensing with robust motion capabilities enables \HandName{} to effectively grasp diverse objects, including challenging cases like crystal balls (\cref{fig:results_tactile}e), while simultaneously capturing detailed contact information (\cref{fig:results_tactile}f). This sensory data enables accurate object pose estimation during manipulation (\cref{fig:results_tactile}g). Additional demonstrations are provided in \href{\suppUrlnmi}{Supplementary Video}.

Through this combination of dense tactile arrays and advanced motor capabilities, \HandName{} achieves unprecedented biomimetic fidelity, advancing both robotic manipulation capabilities and our understanding of human manual dexterity.

\subsection*{Powering \HandName{} with human-like diverse grasping}

While \HandName{'s} high articulation enables sophisticated manipulation, it presents unique challenges in grasp planning. The increased degrees of freedom make traditional mechanical equation-based methods computationally intractable~\cite{siciliano2008springer}. Learning-based alternatives~\cite{ichnowski2020deep}, though avoiding complex analytical solutions, require extensive training data that is both costly to collect and potentially biased by human demonstration preferences---a particular challenge for highly articulated dexterous hands.

\begin{figure}[t!]
    \centering
    \includegraphics[width=\linewidth]{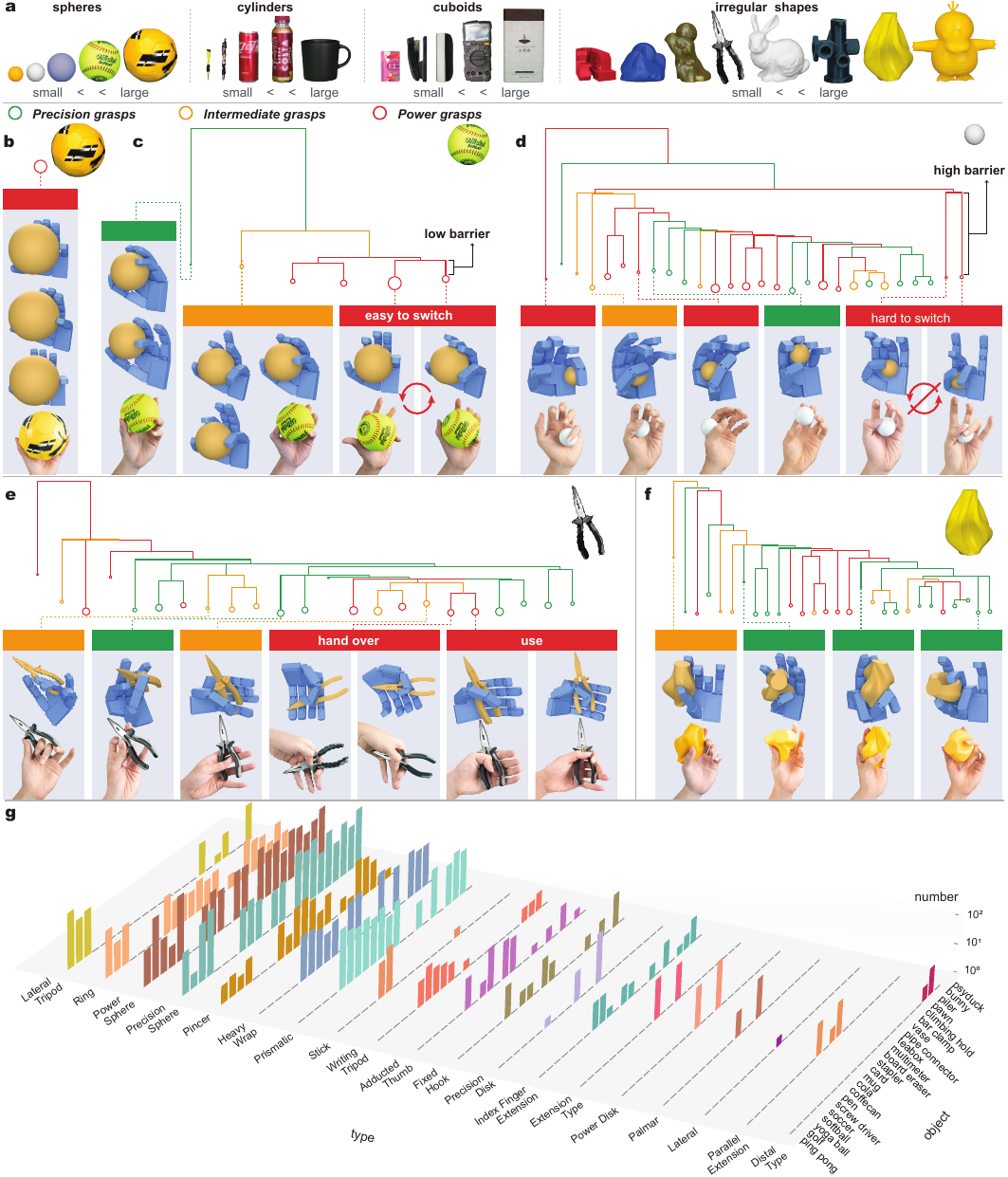}
    \caption{\textbar~\textbf{Human-like diverse grasping strategies for single objects.} \textbf{a}, A total of 23 objects, varying in dimensions and geometrical complexity, are chosen to assess the efficacy of the grasp generation method. \textbf{b--d}, The landscape of possible grasps is depicted as a disconnectivity graph, generated using the \ac{adelm} algorithm. \textbf{e}, For the piler, the diversity in grasp strategies mirrors its common usage in daily human activities. \textbf{f}, This diversity in grasping strategies is maintained even when the object possesses complex geometric features. \textbf{g}, Comprehensive categorization of the generated grasps for all 23 objects, based on human grasp types adopted from~\cite{feix2015grasp}. The results collectively cover all 19 common grasp types, implying the human-like diversity of the generated grasping strategies.}
    \label{fig:results_algorithm}
\end{figure}

We model robotic grasp generation of rigid objects as sampling hand poses from a Gibbs distribution conditioned on object geometry. Each grasp is associated with an energy value derived from force closure criteria, which evaluates how well the grasp can resist external forces. Lower energy values indicate better grasping capability; see `Probabilistic formulation for grasp generation' in \hyperref[sec:methods]{Methods} and Supplementary Information \cref{sec:DFC} for details. Due to the hand's high \acp{dof} and the non-convex nature of the problem~\cite{liu2021synthesizing}, we sample grasps from random initializations and apply a modified \ac{mala} algorithm to reduce energy and escape local minima, converging to low-energy, high-quality grasps; see `Exploration algorithm for complex energy landscape' in \hyperref[sec:methods]{Methods} for details. We validated the approach using a diverse test set of 23 objects, including spheres, cylinders, cuboids, and irregular shapes (\cref{fig:results_algorithm}a). By executing the algorithm from various initializations, the \HandName{'s} biomimetic kinematics (\ref{fig:extended_DH}) and varied object geometries result in diverse grasping poses (\href{\suppUrlnmi}{Supplementary Video}). 

The resulting grasping poses are analyzed through the \ac{adelm} algorithm~\cite{hill2019building} to visualize the complex energy landscape defined in \cref{eq:energy_function}, as shown in \cref{fig:results_algorithm}b--d. In this visualization, circles represent local minima (areas of low energy), where each local minimum contains at least one feasible grasp. The circle's size indicates how many similar grasps exist within that local minimum. The circles are color-coded according to the majority grasp type based on Feix \etal~\cite{feix2015grasp}: \textit{Power} (red), \textit{Precision} (green), and \textit{Intermediate} (yellow). The vertical connecting bars between circles represent energy barriers between different local minima, where shorter bars indicate easier transitions between grasping poses, while longer bars signify transitions that are more difficult to achieve. Direct comparisons between generated grasps and human demonstrations (\cref{fig:results_algorithm}b--d) in the boxes below validate the human-like nature of our solutions. This approach maintains its effectiveness even for challenging cases like pliers and adversarial objects~\cite{mahler2017dex} (\cref{fig:results_algorithm}e--f), with comprehensive energy landscapes presented in \cref{fig:extened_large_landscape}.

To quantitatively assess the human-like diversity of our approach, we analyzed \num{1800} generated grasps according to Feix's taxonomy~\cite{feix2015grasp}, categorizing them into 19 common grasp types; see Supplementary Information \cref{sec:supp:adopted-taxonomy}. The resulting distribution (\cref{fig:results_algorithm}g) demonstrates comprehensive coverage across the human grasp repertoire, from frequent strategies like \textit{Power Sphere} and \textit{Precision Sphere} to specialized configurations such as \textit{Distal Type} and \textit{Palmar} grasps.

Further analysis using contact maps~\cite{li2023gendexgrasp} reveals natural clustering patterns that align with human grasp classifications. By applying dimensionality reduction through \ac{pca} and visualization via \ac{t-SNE} (\cref{fig:extended_algorithm}), we observe distinct groupings of \textit{Power} and \textit{Precision} grasps, with \textit{Intermediate} grasps appropriately positioned near the boundary defined by an \ac{rbf}-kernel \ac{svc}. This distribution mirrors human grasp categorization patterns, where \textit{Intermediate} grasps share characteristics of both primary types (computational details in Supplementary Information \cref{sec:supp:tsne-details}).

The demonstrated ability to generate diverse, human-like hand configurations provides \HandName{} with both optimal and near-optimal control strategies. This algorithmic foundation, working in concert with the low-level controller, enables enhanced dexterity and adaptability in real-world manipulation scenarios.

\subsection*{Adaptive behaviors of \HandName{}}

The integration of advanced tactile sensing with diverse grasping strategies enables \HandName{} to implement a closed-loop sensory-motor feedback mechanism, allowing real-time adaptation to environmental changes. The implementation details are described in `Context-sensitive motor controls' in \hyperref[sec:methods]{Methods}.

\begin{figure}[t!]
    \centering
    \includegraphics[width=0.96\linewidth]{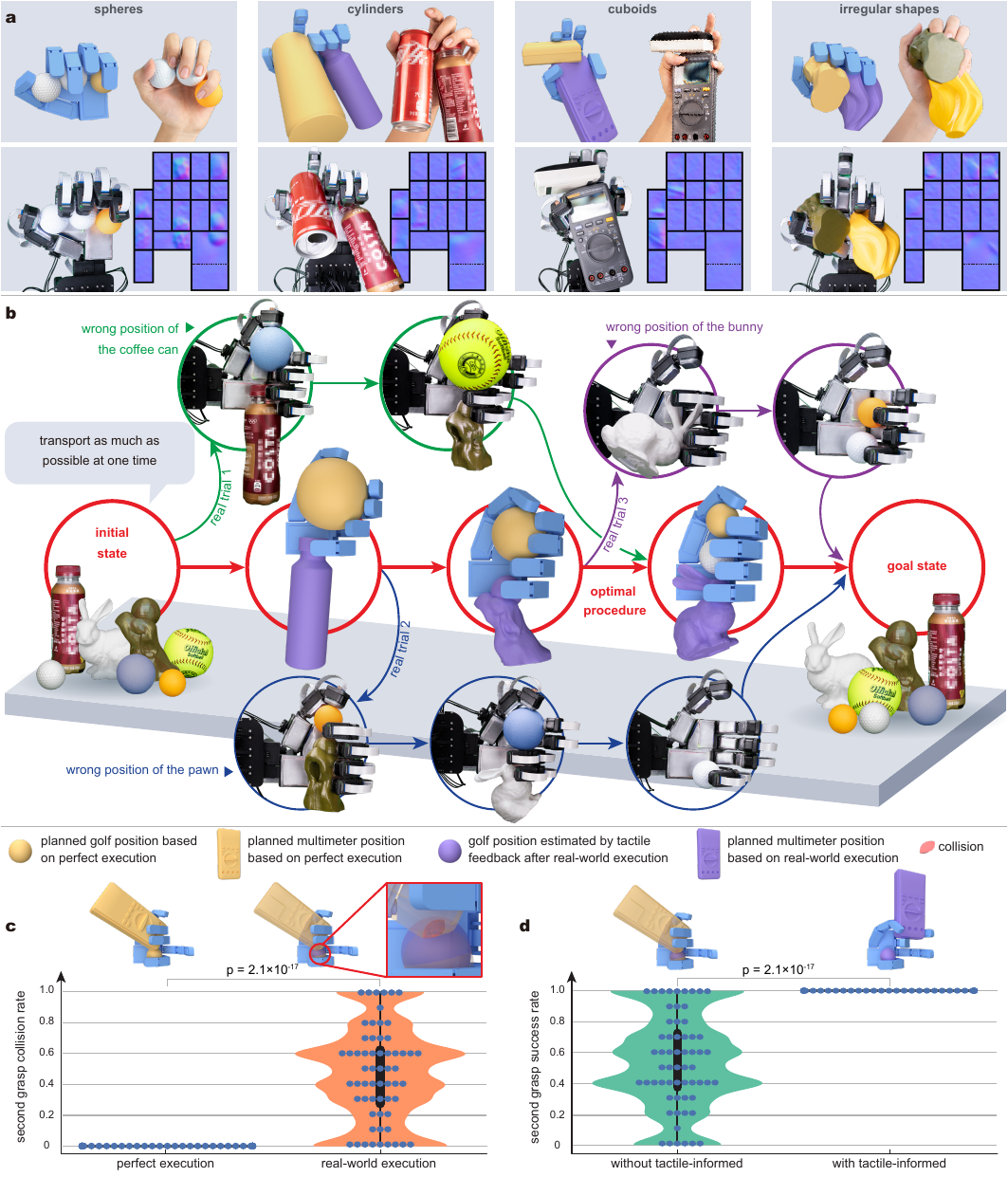}
    \caption{\textbf{\textbar~\HandName{}'s adaptive behaviors.} 
    \textbf{a}, \HandName{}'s ability to grasp multiple objects simultaneously in a human-like manner. \textbf{b}, Despite execution noise, \HandName{} can optimize object transport through multi-object grasping. \textbf{c}, Real-world disturbances significantly increase second grasp collision rates (inset shows collision detection point in red), highlighting the necessity of tactile monitoring. \textbf{d}, Tactile-informed adaptation significantly improves second grasp success rates. Plots \textbf{c} and \textbf{d} show 60 independent technical replicates (blue dots), each representing a unique object combination tested across 10 trials. Box plots show 25th-75th percentiles (box), median (center line), and minimum/maximum values (whiskers). Shaded areas (orange in \textbf{c}, green in \textbf{d}) show kernel density estimations of data distribution. The identical p-values (\(2.1\times10^{-17}\)) displayed on both panels derive from one-sided paired t-tests (\(t(59)=11.8\)) comparing perfect \vs real-world execution (\textbf{c}) and with \vs without tactile-informed adaptation (\textbf{d}). No adjustments were made for multiple comparisons.}
    \label{fig:results_experiments}
\end{figure}

We demonstrate \HandName{'s} capabilities through multi-object grasping~\cite{billard2019trends}, a critical benchmark for hand dexterity that surpasses the limitations of 1-\ac{dof} parallel grippers. This challenging task demands precise contact detection and strategic adjustments to avoid collisions---capabilities that remain elusive for current \ac{ai} systems~\cite{billard2019trends}. While recent advances~\cite{yao2023exploiting,li2024grasp} show promise, managing the stochastic nature of real-world objects, especially those with complex geometries, remains challenging. \HandName{} overcomes these limitations through precise contact point identification (\cref{fig:results_experiments}a).

To evaluate real-world performance, we mounted the \HandName{} on a Kinova Gen3 robotic arm for multi-object transport tasks (\cref{fig:results_experiments}b). The goal is to grasp as many objects as possible in one go to maximize transportation efficiency. While optimal strategies exist under ideal conditions (red route in \cref{fig:results_experiments}b), real-world variables---such as imperfect robot positioning and object perception---require adaptive motor control. The other colored routes in \cref{fig:results_experiments}b illustrate actual scenarios where the \HandName{} encountered wrong object positions but leveraged its comprehensive tactile sensing capabilities (\cref{fig:results_tactile}g) to assess situations and dynamically switch to alternative strategies that accommodate available space, even if theoretically suboptimal. Additional demonstrations of adaptive behaviors, including responses to finger impairments and ball size adaptation, are shown in \ref{fig:extended_flexible_grasp} and \href{\suppUrlnmi}{Supplementary Video}.

We quantified the impact of tactile sensing through extensive experiments involving 60 object combinations across 600 real-world trials. Initial grasps were programmed from a disembodied \ac{ai} perspective, using theoretically optimal strategies without considering environmental dynamics. Each combination underwent 10 real-world trials, with tactile feedback assessing in-hand object positions and potential collision risks. Collision detection involves two steps: using tactile information to estimate the grasped object's pose as shown in \cref{fig:results_tactile}g, and then checking whether the union of the grasped object geometry and the next target object geometry is null. The observed collision rate in real-world execution (\(M = 0.465, SD = 0.306\)) differed significantly from theoretical predictions (\(M = 0.000, SD = 0.000\)), highlighting the substantial gap between simulation and reality, \(t(59) = 11.8, p=2.1\times10^{-17}\) (\cref{fig:results_experiments}c).

\HandName{'s} adaptive capabilities become particularly evident when comparing tactile-informed versus non-tactile-informed control. Upon detecting collision risks, the system rapidly (\({\sim}100~\mathrm{ms}\)) switches to alternative strategies that might be suboptimal in theory but practical in reality.  In scenarios with potential collisions, the non-tactile-informed approach inevitably fails, while the tactile-informed approach maintains productivity through adaptive replanning. The tactile-informed approach achieved perfect adaptation (\(M = 1.000, SD = 0.000\)) compared to significantly lower success rates without tactile feedback (\(M = 0.535, SD = 0.306\)), \(t(59) = 11.8, p=2.1\times10^{-17}\) (\cref{fig:results_experiments}d). Notably, in collision-free scenarios, both approaches demonstrate comparable execution times, with tactile-informed collision checking adding only \({\sim}1~\mathrm{s}\) of processing time, indicating that tactile sensing provides critical robustness with minimal computational overhead during normal operations. The detailed method and the logic chain are available in `Context-sensitive motor controls' in \hyperref[sec:methods]{Methods}.

\clearpage
\section*{Discussion}

\HandName{} represents a significant advance in robotic sensory-motor integration, achieving unprecedented integration of comprehensive tactile sensing with human-like dexterity. Its high-density tactile coverage ($70~\%$ of palmar surface, \(\num{10000}~\mathrm{taxels}/\mathrm{cm}^2\)) substantially exceeds current robotic hand capabilities. This exceptional sensing is achieved through the effective integration of vision-based tactile sensors, physics-based calibration methods, and specialized electronics---all while maintaining full motion capabilities.

Recent advances in tactile sensing~\cite{yousef2011tactile,fishel2012sensing,kim2014stretchable,liu2015finger,yuan2017gelsight,ward2018tactip,li2020f,sun2022soft,li20233,li2024minitac} have primarily focused on parallel grippers. While these sensor-equipped grippers demonstrate enhanced capabilities in specific tasks---such as cable following~\cite{she2021cable}, surface following~\cite{lloyd2024pose}, and articulated object manipulation~\cite{zhao2024tac}---their low-\ac{dof} mechanical structure fundamentally limits their dexterity for complex manipulation.

In contrast, \HandName{'s} integration of comprehensive tactile feedback with high articulation enables more sophisticated manipulation, as demonstrated by its successful multi-object grasping under uncertain conditions. The closed-loop sensory-motor feedback enables context-sensitive adaptations, significantly improving performance in dynamic real-world scenarios. This combination of sensing and adaptability is essential for practical robotics applications requiring safe and efficient environmental interaction.

The design philosophy behind \HandName{} emphasizes replicability, aiming to catalyze broader research in tactile-enabled manipulation. Its achievement of human-like capabilities opens possibilities in prosthetics, teleoperation, collaborative robotics, and human-robot interaction. The hardware's compact, modular architecture facilitates efficient data acquisition and calibration while being adaptable to various robotic platforms. The training-free stochastic optimization approach for grasp generation remains platform-independent, enabling rapid deployment across different hand designs (Extended Data Figure 6). While our current implementation assumes known object geometry, this was a deliberate scope decision to focus on tactile-informed adaptive control rather than geometry reconstruction. Pre-grasping geometry acquisition could be readily integrated using existing vision techniques~\cite{curless1996volumetric,wang2021neus}, and real-time reconstruction during manipulation represents a promising direction for future work. This combination of diverse grasping capabilities and environmental adaptability makes \HandName{} particularly suited for complex manipulation tasks. 

Beyond technical achievements, our results suggest that practical artificial intelligence requires tight integration between sensory processing and strategic adaptation. The demonstrated importance of comprehensive tactile feedback in achieving human-like dexterity aligns with cognitive and neuroscientific perspectives that emphasize the essential role of physical interaction in intelligence~\cite{simon1956rational,merleau19451962,varela2017embodied,vong2024grounded}.

\clearpage
\renewcommand\thesection{Methods}
\section*{Methods}\label{sec:methods}

\subsection*{Tactile sensor fabrication}\label{sec:sensor_design}

The tactile sensor design for \HandName{'s} distal phalanx (\ref{fig:extended_mechatronics_design}a) addresses key challenges in miniaturization and integration. A custom camera module using a single flexible flat cable for both power and data transmission resolves traditional cabling constraints. The sensor housing's U-shaped clevis and tang structure enable the interconnection necessary for anthropomorphic articulation.

Contact detection relies on analyzing elastomer surface deformation through reflected light intensity. To achieve uniform illumination in the confined phalanx space, we developed a specialized Lambertian membrane. This membrane combines an air-brushed spherical aluminum film (mill-resistant matte oil with $1~\mathrm{\upmu m}$ spherical aluminum powder) with a clear silicone base (Smooth-On Inc. Solaris PartA\&B, $1\mathrm{:}1$ ratio). The illumination system comprises surface-mounted LUMILEDS LUXEON 2835 Color Line LEDs (red, green, blue, and white) arranged around an acrylic support, enhanced by light diffuser films. An OV2640 image sensor with $160^\circ$ wide-angle lens provides color-compatible imaging, while $7~\mathrm{mm}\times7~\mathrm{mm}\times4~\mathrm{mm}$ heat sinks ensure thermal stability.

The complete sensing system architecture (\ref{fig:extended_mechatronics_design}b) integrates these components with a custom control module. The module interfaces with cameras through DVP ports, maintaining $240~\mathrm{px}\times240~\mathrm{px}$ image buffers. Spatial resolution achieves $0.1~\mathrm{mm}$ per pixel, verified through known-object calibration. An expanded SPI bus coordinates sequential camera captures, with USB connectivity for PC data transmission and U2D2 protocol for servo control.

The tactile components are integrated into anatomically-scaled phalanx covers matching adult hand dimensions. This modular, single-cable design overcomes traditional challenges in implementing high-resolution, extensive tactile sensing in robotic hands.

\subsection*{\HandName{} fabrication}\label{sec:hand_design}

\HandName{'s} structure (\ref{fig:extended_mechatronics_design}c) integrates 17 compact vision-based sensors in 6 configurations to achieve human-hand proportions. The four fingers---index, middle, ring, and little---share a common architecture (\ref{fig:extended_mechatronics_design}d) with three serial revolute joints: \ac{mcp}, \ac{pip}, and \ac{dip}, each offering $0^{\circ}-90^{\circ}$ range. These joints utilize aluminum shafts supported by deep groove ball bearings, with screw-bushing fixation and torsion springs maintaining a $0^{\circ}$ rest position.

The thumb design (\ref{fig:extended_mechatronics_design}e) features an additional carpometacarpal joint \ac{dof}, enabling $90^{\circ}$ motion range with a $45^{\circ}$ offset from its \ac{pip} joint axis. The two-part palm base facilitates assembly and houses compact tactile sensors, with the upper region incorporating dual cameras in a single sensor for enhanced perception (\ref{fig:extended_mechatronics_design}c).

Finger actuation employs a cable-driven mechanism, with a single cable routed along both sides of each finger's phalanxes, converging at the palm base. Torsion springs facilitate a return to rest position upon cable relaxation. Each finger is powered by a Dynamixel XC330X-T288-T servo motor. For experimental validation, \HandName{} mounts onto a 7-\ac{dof} Kinova Gen3 manipulator.

\subsection*{Probabilistic formulation for grasp generation}\label{sec:grasp_generation_formulation}

The generation of grasp configurations for multi-fingered robotic hands presents significant challenges, particularly when maximizing dexterous capabilities. Instead of relying on data-driven approaches that demand extensive annotated datasets, we formulate grasp generation as a Gibbs distribution sampling problem:
\begin{equation}
    P(H|O) = \frac{1}{Z} \exp^{- E(H,O)}, \label{eq:p}
\end{equation}
where \(H = (T, q)\) represents the hand's pose and joint configurations, \(O\) denotes the target object, \(E(H,O)\) defines the grasping energy function, and \(Z\) is the intractable normalizing constant. The hand's surface geometry \(S(H)\) is computed through forward kinematics.

This energy function combines two weighted components---grasp quality energy \(E_\mathrm{grasp}\) and physical plausibility energy \(E_\mathrm{phy}\):
\begin{equation}
    E(H,O) = \lambda_\mathrm{grasp} E_\mathrm{grasp}(H,O) + \lambda_\mathrm{phy} E_\mathrm{phy}(H,O). \label{eq:energy_function}
\end{equation}

To assess the quality of the grasp, we use force closure criteria to define \(E_\mathrm{grasp}(H,O)\):
\begin{equation}
    E_\mathrm{grasp} = \min_{x \subset S(H)} FC(x, O), \label{eq_grasp_quality}
\end{equation}
where \(x = \{x_i\}\) represents frictional contact points on the hand surface \(S(H)\), and \(FC(x, O)\) assesses force closure formation on the object.

The physical plausibility energy \(E_\mathrm{phy}\) enforces physical constraints by penalizing hand-object penetration and joint limit violations:
\begin{equation}
    E_\mathrm{phy}(H,O) = \sum_{v \in S(H)} \max(-d^\mathrm{SDF}_O(v), 0) + \sum_{j=1}^J \left[ \max (q_j - q^{\max}_j, 0) + \max (q^{\min}_j - q_j, 0) \right] \label{eq:energy_phy_plaus},
\end{equation}
where \(d^\mathrm{SDF}_O(v)\) defines the \ac{sdf} from point \(v\) to object \(O\), and \([q^{\min}_j, q^{\max}_j]\) specifies joint limits for each of the \(J\) joints.

This probabilistic formulation enables scalable generation of diverse, effective grasp configurations.

\subsection*{Exploration algorithm for complex energy landscape} \label{sec:grasp_synthesis_MALA}

The nonlinearity of hand kinematics and contact point selection creates a complex energy landscape for \(E\), making naive gradient-based sampling prone to suboptimal local minima. We address this through a modified \ac{mala} that alternates between contact point sampling and gradient-based pose optimization.

The algorithm initializes with random hand pose \(H\) and contact points \(x \subset S(H)\). Through \(L\) iterations, it updates \(H\) and \(x\) to maximize \(P(H, O)\). Each iteration stochastically chooses between updating the hand pose via Langevin dynamics or replacing a contact point with a uniform sample from the hand surface. Updates undergo Metropolis-Hastings acceptance criteria, favoring lower-energy configurations.

This combination of stochastic updates enables escape from local minima, while Metropolis acceptance guides sampling toward low-energy configurations. An algorithm efficiency analysis is detailed in Supplementary Information \cref{sec:supp:algorithm_detailed_results}.

\subsection*{Context-sensitive motor controls}\label{sec:flexible_grasp_method}

\ref{fig:method_flexible_grasps} demonstrates adaptive control in a four-ball transport scenario, where ball repositories are weighted and combined by volume. Initially, at $t_1$, \HandName{} plans to grasp a golf ball and softball using its little finger and remaining digits. To illustrate the control mechanism, we introduce a manual perturbation during golf ball acquisition, causing \HandName{} to secure the golf ball with its index finger at $t_2$. The occupation of the index finger invalidates the planned softball grasp (light gray in \ref{fig:method_flexible_grasps}), necessitating a strategy revision. Through comprehensive tactile sensing, \HandName{} detects the situation and adapts by executing an alternative approach---grasping a yoga ball using its thumb, index, and middle fingers. While this solution was initially considered suboptimal, it demonstrates the system's capacity for real-time adaptation to unexpected conditions.

\clearpage
\section*{Data Availability}

The data that support the findings of this study are available from  Zenodo~\cite{zhao2025codedata} (\url{https://doi.org/10.5281/zenodo.10141935}). 

\section*{Code Availability}

The code used for performing grasp synthesis and training calibration models is available from  Zenodo~\cite{zhao2025codedata} (\url{https://doi.org/10.5281/zenodo.10141935}). 

\section*{Acknowledgments}

We thank Miss Z. Chen (BIGAI) and Miss Q. Gao (PKU) for their exceptional work on figure design. We are grateful to H. Liang (BIGAI) for his assistance with mechanical design, to Mr. M. Toszeghi (QMUL) for his meticulous proofreading efforts, to Dr. W. Yuan (UIUC), Dr. B. Dai (PKU, BIGAI) and Dr. Y. Su (BIGAI) for engaging in discussions, to Z. Qi (THU) for his assistance with grasping classification data, to Y. Niu (PKU, BIGAI) for his dedication in coding the Kinova driver, to Dr. L. Ruan (UCLA) for his assistance in voiceover, and to Prof. Y. Yang (PKU) and Prof. Y. Wang (PKU) for their contributions to the Shadow Hand hardware, including the video setup by Q. Wang (PKU). We acknowledge the support from J. Cui (BIGAI, PKU), Y. Ma (BIGAI), Y. Wu (BIGAI, PKU), and M. Han (UCLA) in creating portions of our supplementary video. Special thanks are due to Dr. W. Zhang and Dr. L. Li from the 301 Hospital for their professional expertise in human hand tactile sensing, and to Dr. R. Zhang and NVIDIA for their generous GPUs and hardware support. We are immensely grateful to Offbeat Ripple Studio for their invaluable expertise and collaboration in producing the supplementary video. Lastly, we extend our gratitude to the National Comprehensive Experimental Base for Governance of Intelligent Society, Wuhan East Lake High-Tech Development Zone, for their generous support. This work is supported in part by the National Science and Technology Major Project (2022ZD0114900; S.-C.Z. and Y.Z.), the National Natural Science Foundation of China (62376009; Y.Z.), and the Beijing Nova Program (20230484487; Y.Z.).

\section*{Author Contributions Statement}

Z.Z.: building the hardware, devising control algorithms, coding, designing studies, running the \HandName{} adaptive behavior study, analyzing data, and writing.
W.L.: building tactile sensors, coding, running the \HandName{} adaptive behavior study, analyzing data, and writing.
Y.L.: devising grasp synthesis algorithms, analyzing data, running the diverse grasp generation study, organizing data annotation, coding, and writing.
T.L.: devising grasp synthesis algorithms, coding, and writing.
B.L.: devising tactile sensor calibration algorithms, coding, analyzing data, and writing.
M.W.: building the PCB, and writing.
K.D.: brainstorming ideas.
H.L: conceiving and directing the research, and writing.
Y.Z.: brainstorming ideas, writing, directing the research, and providing the environment and funding support for conducting this research.
Q.W.: funding support for conducting this research.
K.A.: brainstorming ideas, directing the work on robotic tactile sensing, and writing.
S.-C.Z.: providing the environment and funding support for conducting this research.

\section*{Competing Interests Statement}

The authors declare that they have no competing interests. 

\setcounter{figure}{0}
\renewcommand{\figurename}{}
\renewcommand\thefigure{Extended Data Figure~\arabic{figure}}
\renewcommand\thetable{Extended Data Table~\arabic{table}}

\clearpage
\begin{figure}[t!]
    \centering
    \includegraphics[width=\linewidth]{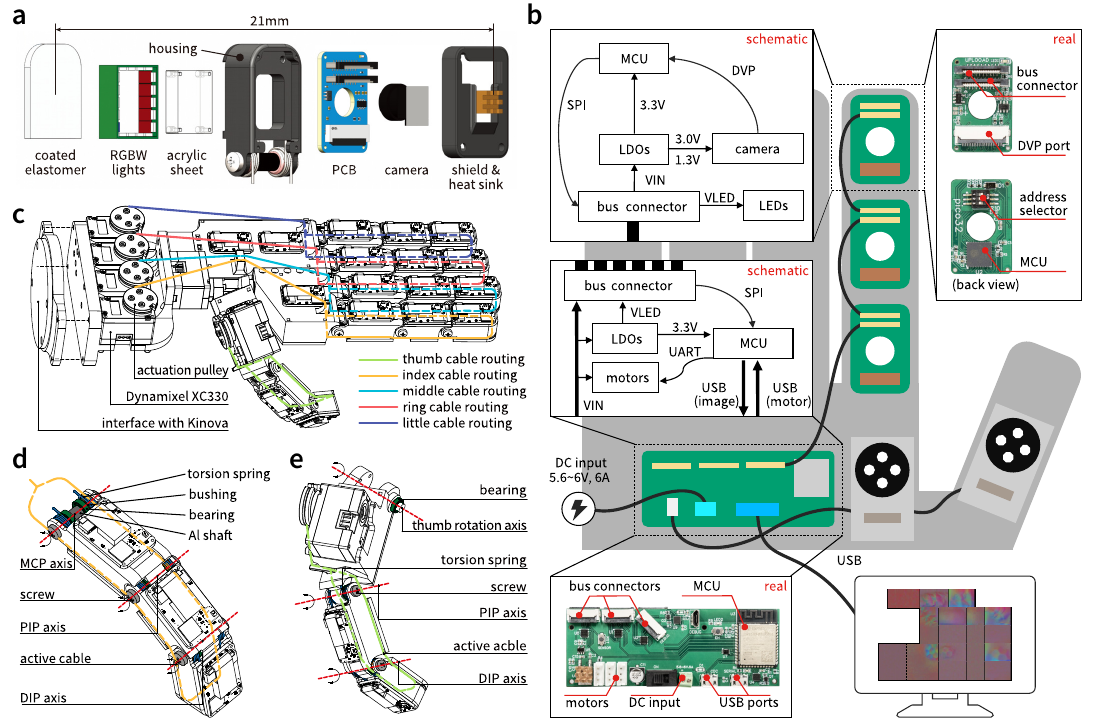}
    \caption{\textbf{\textbar~Mechatronic design of \HandName{}.} \textbf{a}, Exploded view of a vision-based tactile sensor as a distal phalanx. \textbf{b}, Electrical components and system scheme. \textbf{c}, Schematic of \HandName{} assembly and cable-driven mechanism. \textbf{d}, Finger model with mechanical components. \textbf{e}, Thumb model with mechanical components.}
    \label{fig:extended_mechatronics_design}
\end{figure}

\clearpage
\begin{figure}[t!]
    \centering
    \includegraphics[width=\linewidth]{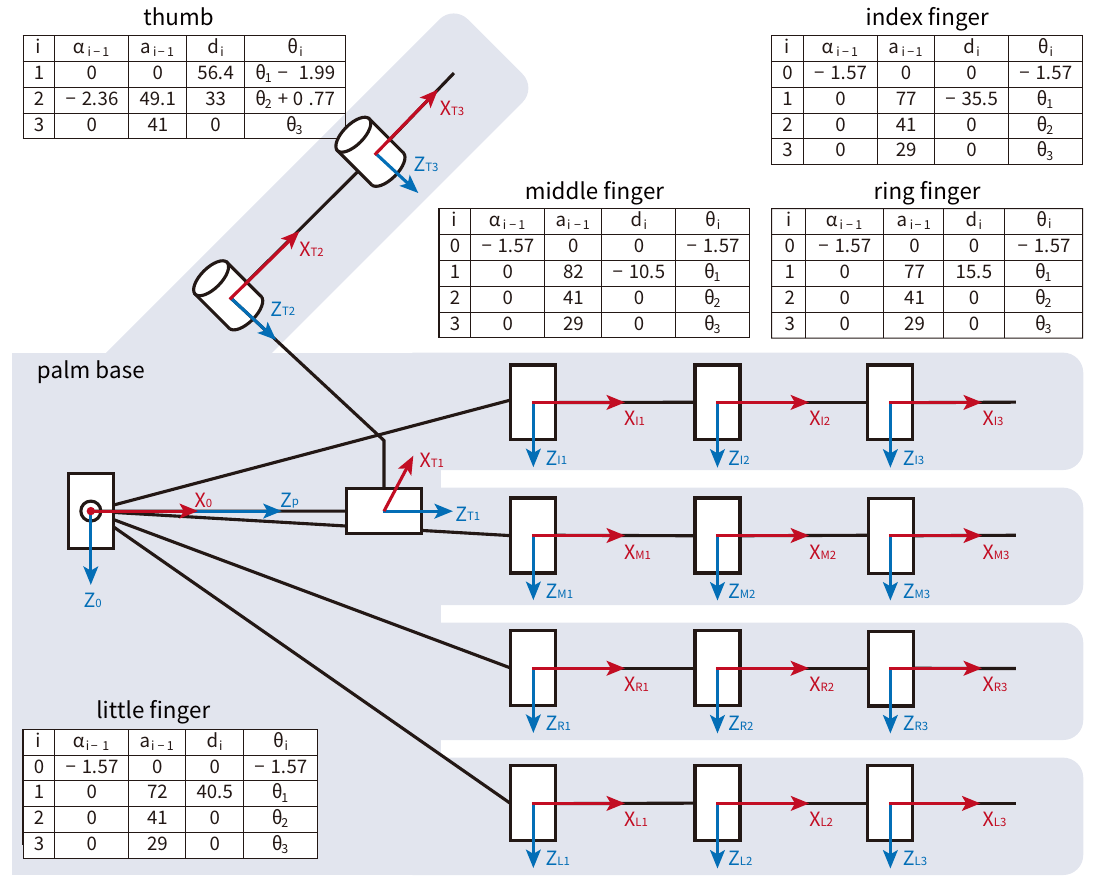}
    \caption{\textbf{\textbar~Kinematic model of the \HandName{}.} We adopt the modified Denavit-Hartenberg (DH) norm to establish the coordinates for the palm base and finger phalanxes. The transformations between these coordinates are represented in DH tables. In these tables, $a_{i-1}$ is the distance along ${X}_{i-1}$ between ${Z}_{i-1}$ and ${Z}_{i}$, $\alpha_{i-1}$ is the angle about ${X}_{i-1}$ between ${Z}_{i-1}$ and ${Z}_{i}$, $d_i$ is the distance along ${Z}_{i}$ between ${X}_{i-1}$ and ${X}_{i}$, and $\theta_i$ is the angle about ${X}_{i}$ between ${X}_{i-1}$ and ${X}_{i}$.}
    \label{fig:extended_DH}
\end{figure}

\clearpage
\begin{figure}[t!]
    \centering
    \includegraphics[width=\linewidth]{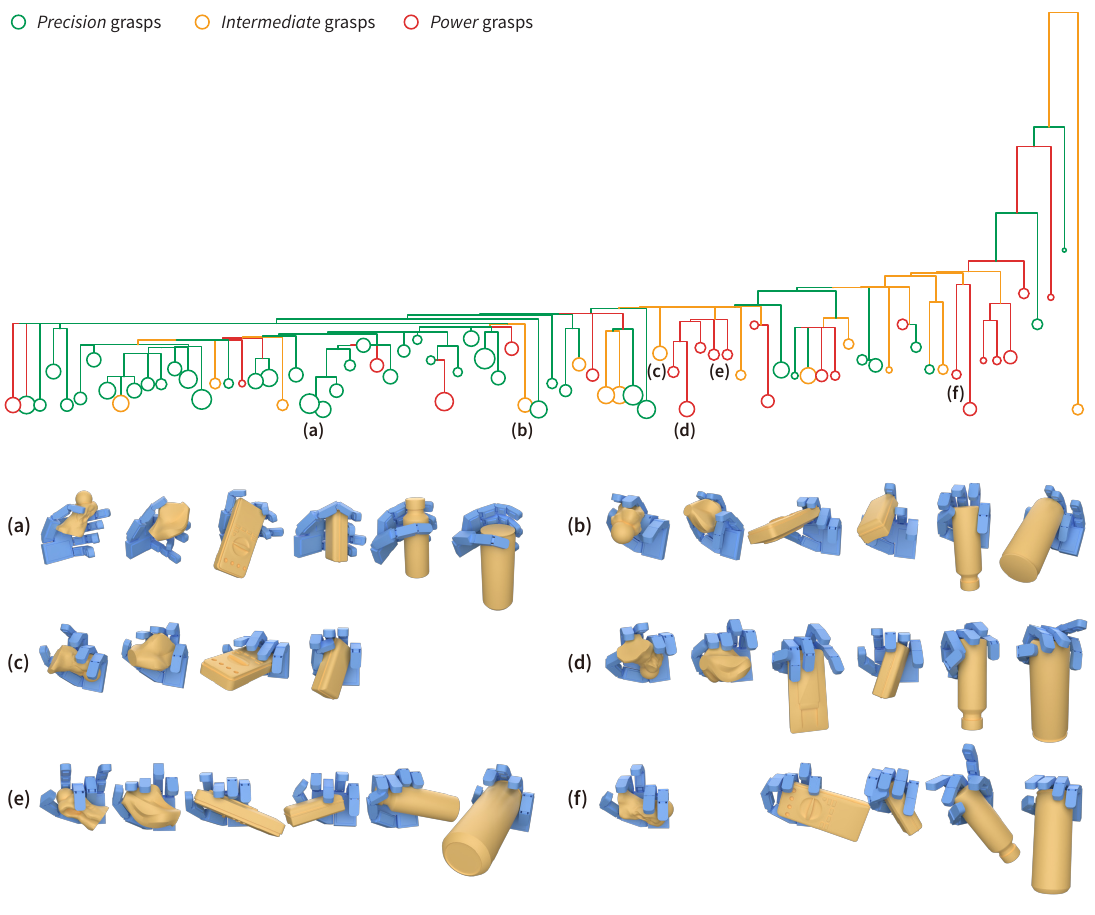}
    \caption{\textbf{\textbar~Multi-object landscape.} We examine the grasp relationships among six example objects (pawn, vase, multimeter, board eraser, coffee bottle, and Coke can) using a large disconnectivity graph. This landscape comprises 79 basins, each categorized into one of three grasp types (\textit{Power}, \textit{Intermediate}, \textit{Precision}) based on the majority of grasps it contains. Within each basin, similar grasp strategies are observed across different objects, as illustrated in (a)-(f).}
    \label{fig:extened_large_landscape}
\end{figure}

\begin{figure}[t!]
    \centering
    \includegraphics[width=\linewidth]{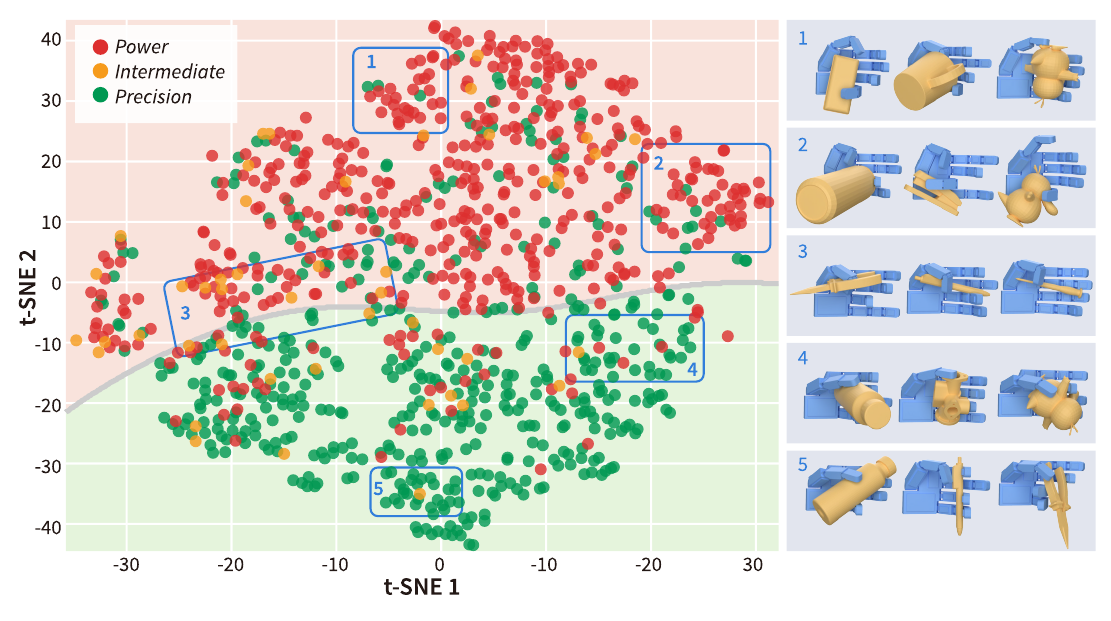}
    \caption{\textbf{\textbar~Extended results of the grasp generation algorithm.} Visualization of grasp samples with \ac{t-SNE} reveals that most \textit{Power} grasps and \textit{Precision} grasps are clustered separately, with \textit{Intermediate} grasps lying in between. This map indicates a strong alignment between the generated results and human definitions of grasp types.}
    \label{fig:extended_algorithm}
\end{figure}

\clearpage
\begin{figure}[t!]
    \centering
    \includegraphics[width=\linewidth]{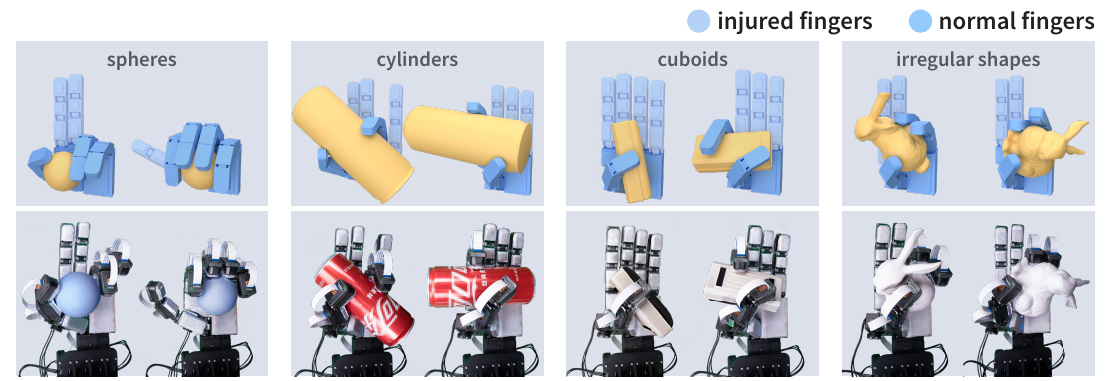}
    \caption{\textbf{\textbar~More adaptive behaviors by the \HandName{}.} \HandName{}'s stable grasping with some fingers disabled (shown in light gray), mirroring human compensation for finger injuries.}
    \label{fig:extended_flexible_grasp}
\end{figure}

\clearpage
\begin{figure}[t!]
    \centering
    \includegraphics[width=\linewidth]{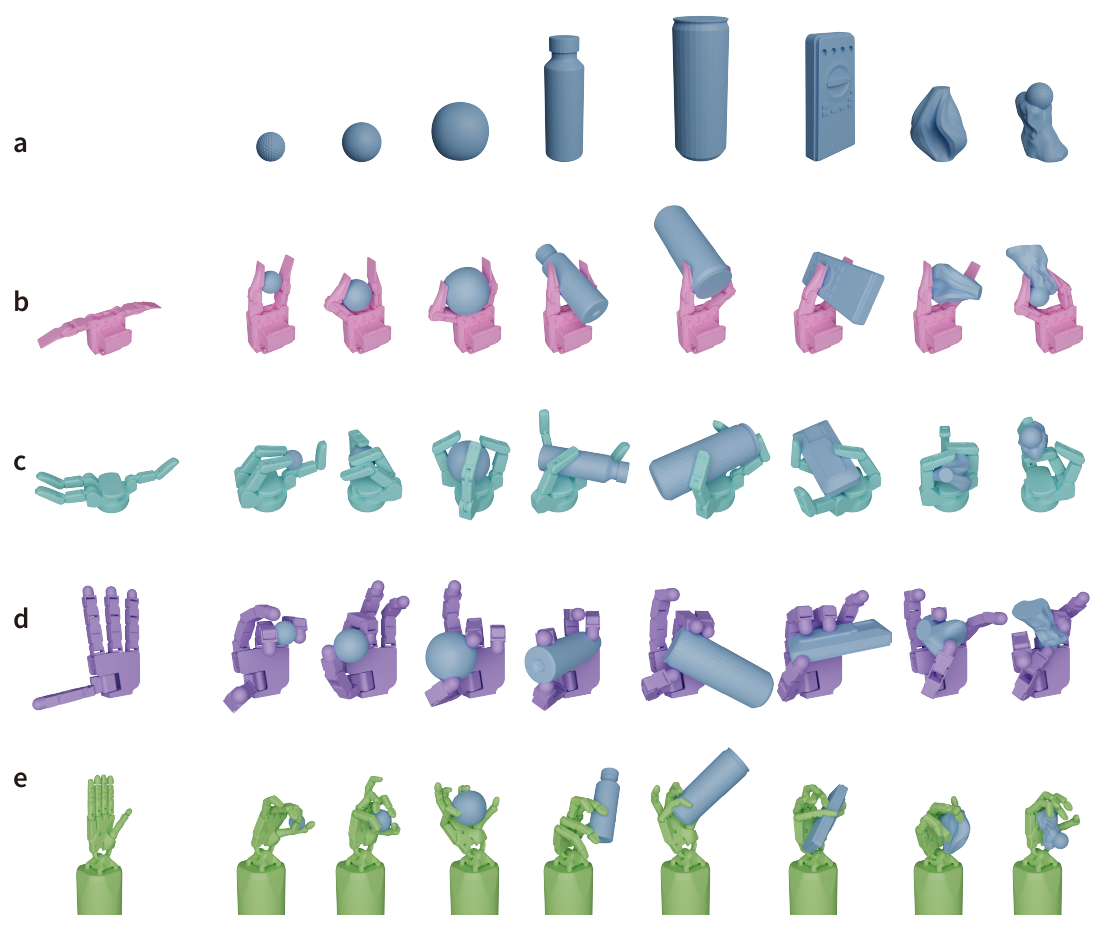}
    \caption{\textbf{\textbar~Extension to other hand topologies.} Our algorithm generalizes to various hand types without requiring specific mechanical structures or training samples. \textbf{a}, eight objects are used for testing with four different hands: \textbf{b}, two-finger EZGripeer~\cite{EZGripper}, \textbf{c}, Barrett 3-fingered Gripper~\cite{Barrett}, \textbf{d} four-finger Allegro Hand~\cite{Allegro}, and \textbf{e}, anthropomorphic Shadow Hand~\cite{shadowrobot}.}
    \label{fig:extended_other_hands}
\end{figure}

\clearpage
\begin{figure}[t!]
    \centering
    \includegraphics[width=\linewidth]{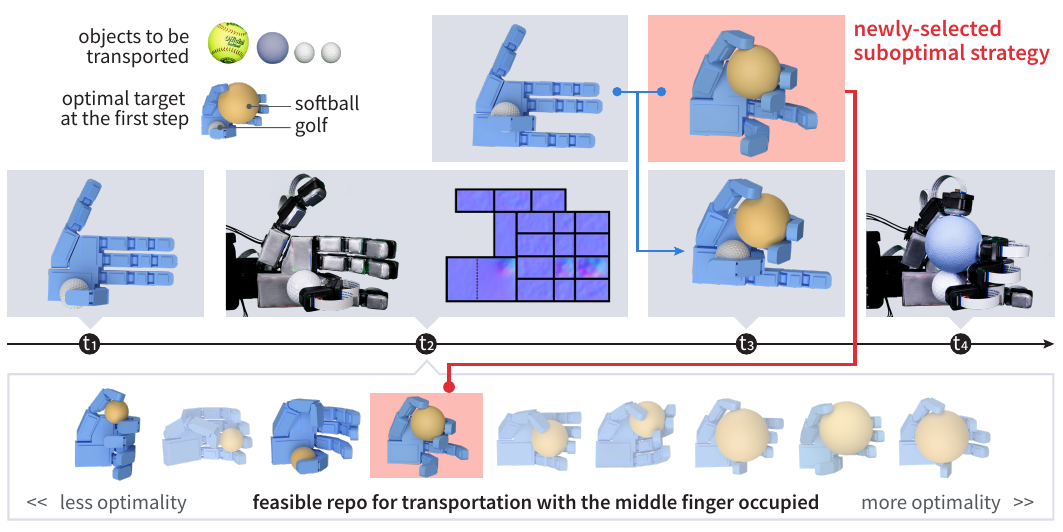}
    \caption{\textbf{\textbar~Illustration of methods to realize human-like dexterous grasping.} \HandName{} employs a two-stage strategy for multi-object transportation. It adjusts for in-hand position variations due to perturbations, dynamically adapting its second-stage strategy to prevent collisions and maximize efficiency (Light gray in the grasping repository indicates that grasping strategies are rendered infeasible at the current time).}
    \label{fig:method_flexible_grasps}
\end{figure}

\clearpage
\bibliography{reference_header,reference}

\clearpage
\setcounter{page}{1}
\setcounter{section}{0}
\setcounter{figure}{0}
\setcounter{equation}{0}
\setcounter{table}{0}
\renewcommand{\figurename}{}
\renewcommand\thefigure{Supplementary Figure~\arabic{figure}}
\renewcommand\thetable{Supplementary Table~\arabic{table}}
\renewcommand\thepage{S\arabic{page}}
\renewcommand\thesection{S\arabic{section}}
\renewcommand\theequation{S\arabic{equation}}

\section*{Supplementary information}

{\normalsize{\textbf{\TitleSupp}}

\noindent{}Zihang Zhao$^{1,2\,\dagger}$, Wanlin Li$^{2\,\dagger}$, Yuyang Li$^{1,2\,\dagger}$, Tengyu Liu$^{2\,\dagger}$, Boren Li$^{2}$, Meng Wang$^{2}$, Kai Du$^{1}$, Hangxin Liu$^{2\,*}$, Yixin Zhu$^{1,3\,*}$, Qining Wang$^{4}$, Kaspar Althoefer$^{5\,*}$, Song-Chun Zhu$^{1,2}$\\
$\dagger$ equal contributors\quad{}$^*$ corresponding authors\\
$^1$ Institute for Artificial Intelligence, Peking University, Beijing 100871, China\\
$^2$ Beijing Institute for General Artificial Intelligence (BIGAI), Beijing 100080, China\\
$^3$ PKU-WUHAN Institute for Artificial Intelligence, Wuhan 430075, China \\
$^4$ College of Engineering, Peking University, Beijing 100871, China\\
$^5$ School of Engineering and Materials Science, Queen Mary University of London, London, E1 4NS, UK
}

\clearpage
\section{Sensor characteristics}\label{sec:supp:sensor_characteristics}

\begin{figure}[t!]
    \centering
    \includegraphics[width=\linewidth]{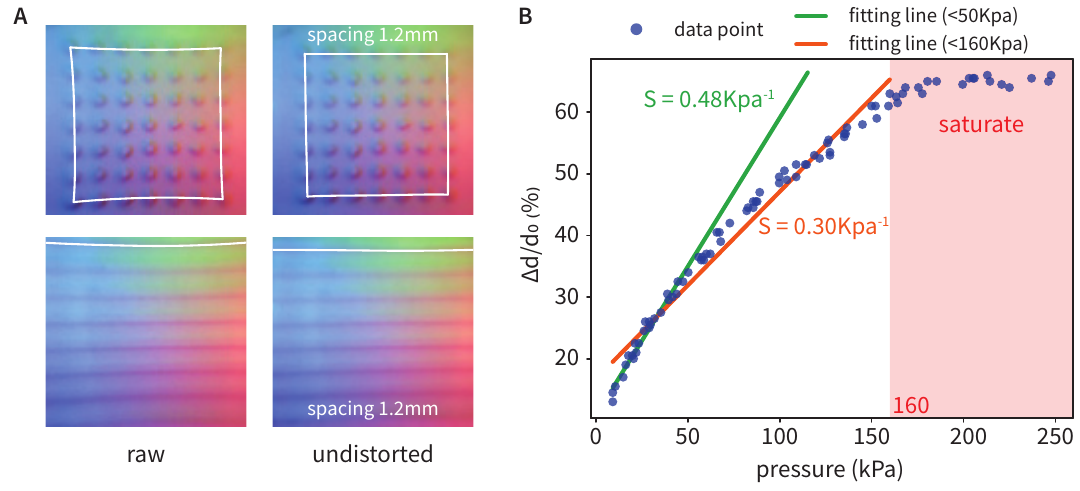}
    \caption{\textbf{\textbar~Detailed sensor characteristics and performance metrics.} (A) The embedded camera undergoes calibration to eliminate distortion using Zhang's method~\cite{zhang2002flexible}. Through controlled testing with objects of precise dimensions, we establish that \(1.0~\mathrm{mm}\) corresponds to \(10\) pixels (taxels) on average, enabling accurate spatial representation. (B) The sensor demonstrates consistent linear response (\(R^2\) = \(0.967\)) with a sensitivity of \(0.30~\mathrm{kPa}^{-1}\) up to its saturation point at \(160~\mathrm{kPa}\). For typical human-like grasping tasks (\(\sim20~\mathrm{kPa}\)) as quantified in research by Jiang \etal~\cite{jiang2024capturing}, the sensor exhibits an enhanced sensitivity of \(0.48~\mathrm{kPa}^{-1}\) at pressures below \(50~\mathrm{kPa}\) (\(R^2\) = \(0.973\)). In the figure, $d_0$ represents the initial thickness of the elastomer in the absence of external forces, and $\Delta d$ denotes the absolute change in thickness when the elastomer is subjected to external forces.}
    \label{fig:sensor_characteristics}
\end{figure}

We estimate the spatial resolution of the sensing using the following procedures. Due to inherent optical distortions in real-world camera systems, particularly pronounced at the edges of the field of view (\ref{fig:sensor_characteristics}A), we first implemented a comprehensive camera calibration protocol. Following Zhang's method~\cite{zhang2002flexible}, we utilized a checkerboard calibration pattern to develop a precise distortion correction model. This model effectively transforms distorted image coordinates into their corrected counterparts, ensuring accurate pixel mapping across the entire sensor surface.

The application of the distortion correction model resulted in a uniform pixel-to-taxel correspondence across the corrected field of view. This correction is crucial for maintaining consistent spatial measurements and ensuring the reliability of tactile sensing data across the entire sensor surface. To quantify the spatial resolution of our tactile sensor system, we conducted systematic measurements using test objects with precisely known dimensions. Through experimental validation, we established that $1.0~\text{mm}$ of physical distance corresponds to $10$ pixels (taxels) in the sensor output. This calibration yields a spatial resolution of $0.1$ mm per pixel.

The sensor characterization was conducted using a precision measurement setup consisting of a spherical probe with a radius of $1.5~\mathrm{mm}$. The probe was mounted with an ATI force/torque sensor to precisely measure the applied normal force. During the evaluation, the probe was pressed against the tactile sensor's surface in the normal direction to assess three key performance metrics: range, linearity, and sensitivity.

As shown in \ref{fig:sensor_characteristics}B, the experimental results demonstrate excellent sensor performance characteristics. The sensor exhibits a strong linear relationship between applied force and output signal, achieving an $R^2$ value of $0.967$ through linear fitting analysis. The sensor maintains reliable operation up to a saturation point of approximately $160~\mathrm{kPa}$, beyond which some performance degradation is observed. Throughout the operational range below the saturation point, the sensor maintains a sensitivity of $0.30~\mathrm{kPa}^{-1}$.

The sensor performs particularly well in the regime relevant to daily object manipulation tasks ($\sim$$20~\mathrm{kPa}$)~\cite{jiang2024capturing}. In the low-pressure range (below $50~\mathrm{kPa}$), both linearity ($R^2 = 0.973$) and sensitivity ($0.48~\mathrm{kPa}^{-1}$) characteristics are further enhanced, making the sensor especially suitable for precision grasping applications. The consistent precision across the operating range, combined with well-defined performance boundaries, demonstrates the sensor's reliability for practical applications.
\clearpage

\section{Details of image formation model}\label{sec:supp:image_formation}

\subsection{Near-field camera model and calibration}

The near-field camera model consists of two main components: camera geometry and radiometry. The geometry employs a perspective projection model~\cite{hartley2003multiple}, suitable for the compact GelSight-inspired sensors where elastomer deformation is significant relative to the camera's working distance. A 3D surface point, ${}^{\{C\}}\widetilde{\mathbf{X}}$, is projected to a 2D image point, ${}^{\{P\}}\mathbf{x}$, using the equation\
\begin{equation}
    {}^{\{P\}}\mathbf{x} = \frac{1}{{}^{\{C\}}Z}
	\mathbf{K}
	{}^{\{C\}}\widetilde{\mathbf{X}},
\end{equation}{}
where $\{C\}$ and $\{P\}$ are the camera and pixel coordinate frames, respectively. The camera intrinsics, $\mathbf{K}$, determine the unit viewer direction, $\bar{\mathbf{v}}$, at each pixel. The world coordinate frame, $\{W\}$, aligns its XY plane with the elastomer support plane and is related to $\{C\}$ through a 3D rigid transformation, serving as camera extrinsics. The camera is calibrated prior to elastomer attachment using Zhang \etal~\cite{zhang2002flexible} to correct lens distortions and obtain intrinsics and extrinsics. Given a uniform and known elastomer thickness, the contact plane equation is straightforwardly derived.

Assuming vignetting effects are negligible due to the central location of the perceptible region in the camera's field of view, the camera radiometry focuses solely on photometric response. Fixed white balance gain is set for the color cameras, and each color channel, $\{F_c\}_{c=R,G,B}$, has its own monotonic photometric response correlating image irradiance, $\{I_c\}_{c=R,G,B}$, to measured intensity, $\{M_c\}_{c=R,G,B}$, as $M_c=F_c(I_c)$. This response is calibrated using $\{G_c=F_c^{-1}\}_{c=R,G,B}$, following Mitsunaga~\etal\cite{mitsunaga1999radiometric}, where $G_c$ is modeled as a third-order polynomial. Surface radiance, $\{L_c^{i,j}\}_{c=R,G,B}$, is then derived from measured intensity as $L_c^{i,j}=G_c\left(M_c^{i,j}\right)$, where $(i, j)$ denotes the pixel location.

Given that all compact GelSight-inspired sensors utilize the same camera type, a single calibration suffices for all, with the exception of an additional calibration image required to estimate the extrinsics for each individual sensor.

\subsection{Ground-truth elastomer surface geometry acquisition}

To construct the sensor model, acquiring ground-truth elastomer surface geometry is essential. Direct measurement using high-accuracy devices being impractical, an indirect approach is adopted. The experimental setup employs an XYZ 3-axis linear trimming stage, as depicted in \ref{fig:simulator}A. The sensor is affixed at the stage's bottom, allowing lateral adjustments in the X and Y directions. Assuming ideal manufacturing, the stage's XY plane is parallel to the sensor contact plane. A 3D-printed cube probe is mounted upside down at the top, perpendicular to the sensor contact plane. The stage coordinate frame, $\{S\}$, aligns its XY plane with the mount's flat base and its Z-axis perpendicular to the sensor contact plane. The ${}^{\{S\}}Z=0$ is set by lowering the cube probe until just before contact. The cube probe's position is adjusted to appear near the perceptible region's center. The 3D transformation from $\{S\}$ to $\{C\}$ is calibrated using a perspective-n-point problem~\cite{lepetit2009ep}, involving camera intrinsics and 3D corner points of the cube probe. Calibration performance is assessed by reprojection error, as shown in \ref{fig:simulator}A.

Upon calibrating the stage, the cube probe can be substituted with any 3D-printed object of known geometry. A depth map is generated using the calibrated camera's visibility constraint. Only pixels with depth within the elastomer thickness range are preserved in the contact area, while the contact-free area is filled using the contact plane's depth map. Thus, by mounting various objects and adjusting the calibrated stage, per-pixel elastomer surface geometry is obtained as ground truth.

\subsection{Near-field lighting model and calibration}

The near-field lighting model comprises lighting geometry and radiometry. In the compact GelSight-inspired sensors, the geometry involves multiple LEDs illuminating the surface from various positions. Given their small size relative to working distance, LEDs are modeled as point sources. Their positions are denoted as $\{{}^{\{C\}}\widetilde{\mathbf{S}}_k\}_{k\in[1, N]}$, where $N$ is the LED count for a sensor. For each $k^{th}$ LED, the unit light direction, $\bar{\mathbf{s}}_k$, and light-surface distance, $\|\mathbf{s}_k\|$, are determined. A 3D-printed calibration object with two cylindrical sundials on a square mount's diagonal is used for light position estimation. The object is carefully pressed onto the elastomer using the calibrated XYZ stage. Individual LEDs are activated sequentially, and their positions are estimated using shadow cues~\cite{sato2003illumination} and triangulation.

Lighting radiometry accounts for the spatial distribution of emitted energy reaching the surface. It is influenced by the LED's radiation pattern and its distance to the surface point. For the chosen LEDs with symmetrical radiation patterns, the radiometry is characterized by
\begin{equation}
    \kappa_{k}^{i,j} = \eta_{k}\left(\left(-\bar{\mathbf{s}}_k^{i,j}\right)^\top\bar{\mathbf{p}}_k\right)^{\mu_k}/\|\mathbf{s}_k^{i,j}\|^2,
\end{equation}
where $\bar{\mathbf{p}}_k$ is the principal light direction for the $k^{\text{th}}$ LED. Parameters $\eta_{k}$, $\mu_k$, and $\bar{\mathbf{p}}_k$ are calibrated jointly with reflectance parameters.

\subsection{Reflectance model and calibration}

The reflectance model for our compact GelSight-inspired sensors is based on the elastomer material's reflectance property. Although the elastomer is theoretically near-Lambertian, its appearance in camera images is flatter, which we attribute to surface roughness. To address this, we adopt the generalized Lambertian model~\cite{nayar1995visual}. The surface radiance under specific lighting conditions is modeled as
\begin{equation}
    \begin{cases}\label{eq:brdf}
        L_c^{i,j}\left(\theta_{r}^{i,j}, \theta_{i}^{i,j}, \phi_{r}^{i,j}-\phi_{i}^{i,j};\rho_{c},\sigma\right) = \kappa^{i,j}\frac{\rho_c}{\pi}\max\left[0,\cos{\theta_i^{i,j}}\right]f_r^{i,j}\\
        f_r^{i,j} = A + B\max\left[0, \cos{\left(\phi_r^{i,j}-\phi_i^{i,j}\right)}\right]\sin\alpha^{i,j}\tan\beta^{i,j}\\
        A = 1.0 - 0.5\frac{\sigma^2}{\sigma^2+0.33}\\
        B = 0.45\frac{\sigma^2}{\sigma^2+0.09} 
    \end{cases},
\end{equation}
where \(\left(\theta_i^{i,j},\phi_i^{i,j}\right)\) and \(\left(\theta_r^{i,j},\phi_r^{i,j}\right)\) are the light and viewer directions in a local coordinate frame. Additional parameters include \(\alpha^{i,j}\), \(\beta^{i,j}\), \(\sigma\), and \(\{\rho_c\}_{c=R,G,B}\).

For calibration, a 3D-printed sphere with a $3~\mathrm{mm}$ diameter is pressed at \(Q\) different locations onto the elastomer surface using a calibrated XYZ stage. At each press, \(N\) images are captured, each with a single LED lit. The calibration aims to solve the following constrained nonlinear fitting problem:
\begin{equation}
    \begin{aligned}\label{eq:calib_cost_fun}
    \min\limits_{\eta_k, \mu_k, \mathbf{\bar{p}}_k, \sigma, \rho_c} &\sum\limits_{q=1}^{Q}\sum\limits_{k=1}^{N}\sum\limits_{c=1}^3\sum\limits_{i=1}^H\sum\limits_{j=1}^W\left[G_c\left(M_{c,k, q}^{i,j*}\right) - \kappa_k^{i,j}\frac{\rho_c}{\pi}\cos\theta_i^{i,j}f_r^{i,j}\right]^2
        + \lambda\sum\limits_{k=1}^N\left(\eta_k - 1\right)^2\\
    \text{s.t.} \hspace{1.3em} & \begin{array}{l@{\quad}l@{}l@{\quad}l}
     & i, j\in\mathcal{I}_k\cap\mathcal{S}\\
         &\|\mathbf{\bar{p}}_k\|=1\\
         &\mu_k\geq0\\
         &\sigma\geq0\\\end{array}
    \end{aligned},
\end{equation}
where \(H\) and \(W\) are the height and width of the image sensor in pixels. \(\mathcal{I}_k\) is the non-shadow area of the \(k^{th}\) image, and \(\mathcal{S}\) is the perceptible region of GelSight-inspired within the camera's field of view. The asterisk denotes measurement throughout the paper. The first term in the equation is for least-square fitting, while the second term is for regularization. Due to the problem's high nonlinearity, careful initialization is required. \(\mathbf{\bar{p}}_k\) is initialized to point towards the center of the perceptible region, \(\mu_k\) is initialized to $1.2$, \(\eta_k\) and \(\rho_c\) are initialized to $1$, and \(\sigma\) is initialized to $0.2$. The problem is solved using the Levenburg-Marquardt algorithm.  Calibration performance is evaluated using a test image not seen during fitting, with the error converging at 4 presses, as shown in \ref{fig:simulator}B. This implies that only $4~\mathrm{N}$ calibration images are needed for accurate parameter estimation. Since the sensors share the same type of LEDs and elastomer material, calibration is required only once.

Finally, with all LEDs turned on, the overall surface radiance, \(\widehat{L}_c^{i,j}\), is given by
\begin{equation}
    \widehat{L}_c^{i,j}=\sum_{k=1}^NL_{c,k}^{i,j}.
\end{equation}

\subsection{Cast shadow model}

In addition to the pixel-wise shading model, global effects like inter-reflection and cast shadow also influence pixel intensities in tactile images. Inter-reflection is negligible due to the black low-albedo matte material of the sensor's cover. For cast shadows, given the known calibrated point light positions and surface geometry, we compute the shadow for each light using the canonical hidden point removal operator~\cite{katz2007direct} to mask the shadow-casting areas.

\subsection{Soft elastomer deformation model}

The contact-free elastomer is assumed to have a 3D geometry parallel to the reference plane. Given the camera extrinsics and elastomer thickness, the contact plane equation in \(\{C\}\) is determined. When an object contacts the elastomer, its depth map is computed using camera visibility constraints, object mesh, and 6D pose. Only pixels within the elastomer thickness range are preserved in the contact area. The non-contact area is filled with the depth map of the contact-free elastomer. Due to the material's softness, the elastomer deforms into a smoothed shape of the contacting object. We apply a simple soft body simulation approximation, smoothing the depth map boundaries between contact and non-contact areas using pyramid Gaussian kernels~\cite{si2022taxim}. Finally, pixel values in non-contact and non-shadow areas are replaced by their counterparts in a background image captured without elastomer deformation.

\subsection{Sensor simulator evaluation}

\begin{figure}[t!]
    \centering
    \includegraphics[width=\linewidth]{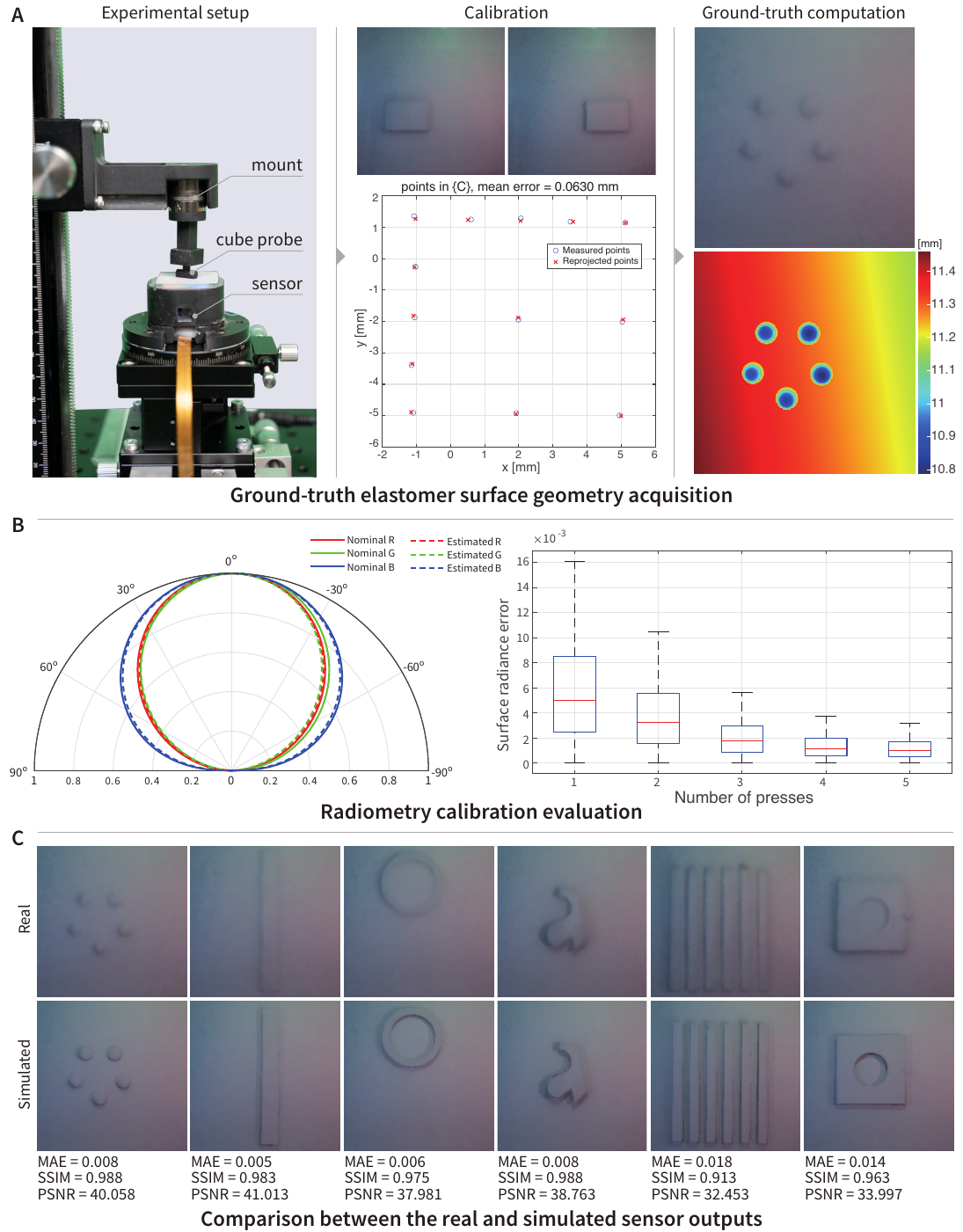}
    \caption{\textbf{\textbar~Sensor calibration and simulation.} \textbf{(A)} Setup for obtaining ground-truth elastomer surface geometry. \textbf{(B)} Calibrated light radial attenuation pattern versus nominal values, evaluated by surface radiance error across different press counts. \textbf{(C)} Real versus simulated sensor outputs with similarity scores indicated.}
    \label{fig:simulator}
\end{figure}

We evaluate the sensor simulator by comparing its outputs to real sensor data, using the setup shown in \ref{fig:simulator}A. We 3D print 20 objects from the tactile shape dataset~\cite{gomes2021generation}, excluding the \textit{Cone} to prevent sensor damage. For each object, we collect multiple sensor outputs under varying conditions, resulting in a dataset of 140 real images.

We employ three metrics for comparison: mean absolute error (L1), structural index similarity (SSIM), and peak signal-to-noise ratio (PSNR). Qualitative and quantitative comparisons are presented in \ref{fig:simulator}C and \ref{tab:quant_gelsim}, respectively. The simulator excels in structural accuracy (SSIM) due to our geometrical calibration method and also accurately replicates image intensity (L1 and PSNR) through lighting radiometry and reflectance calibration. The simulated and real cast shadow regions align well, confirming the effectiveness of our light position calibration.

Our physics-based sensor model and calibration techniques enable the simulator to generate accurate sensor outputs at scale. This eliminates the need for extensive data acquisition for each sensor on \HandName{}. The calibrated parameters can be reused as the hardware components are shared, simplifying the calibration process for all sensors on \HandName{}.

\clearpage
\section{Details of DPS}\label{sec:supp:network}

\subsection{Training details}

We implemented the deep PS network using PyTorch and employed additive white Gaussian noise (standard deviation $0.01$) for data augmentation. The network was trained using the Adam optimizer with $\beta_1=0.9$ and $\beta_2=0.999$, a batch size of 128, and an initial learning rate of $0.01$. The learning rate was halved every five epochs. The training was conducted on a single NVIDIA RTX 3090 GPU and took approximately six hours to converge.

\subsection{Evaluator}

We evaluate the trained deep PS network using an unobserved dataset of four shapes with 46,656 samples. The mean angular error (MAE) serves as the evaluation metric.

\clearpage
\section{Differentiable force closure estimator}\label{sec:DFC}

Determining force-closure grasps under kinematic constraints is computationally expensive in our context. To mitigate this, we introduce a quick, differentiable force closure estimator, facilitating efficient grasp generation. Computational time benchmarks can be found in Supplementary Information \ref{sec:supp:algorithm_detailed_results}.

For a grasp with a set of \(n\) contact points \( \{x_i \in \mathbb{R}^3, i = 1, \ldots, n\} \), it is in force closure if for any external wrench \( \omega \), there exists a combination of contact forces \( \{f_i \in \mathbb{R}^3\} \) that can resist \( \omega \). Specifically, the \(i\)-th contact force \( f_i \) should lie within the friction cone at the \(i\)-th contact point \( x_i \). Formally, a grasp is in force closure if it satisfies the following constraints:
\begin{subequations}
    \begin{align}
        GG' &\succeq \epsilon I_{6 \times 6}, \label{eq_c1} \\
        Gf &= 0, \label{eq_c2} \\
        f_i^T c_i &> \frac{1}{\sqrt{\mu^2 + 1}} |f_i|, \label{eq_c3} \\
        x_i &\in S(O), \label{eq_c4}
    \end{align}
    \label{eq_c}%
\end{subequations}%
where $S(O)$ is the object surface, $c_i$ the friction cone axis at $x_i$, $\mu$ the friction coefficient, $f=[f_1^T f_2^T ... f_n^T]^T\in\mathbb{R}^{3n}$ the unknown variable of contact forces, and
\begin{align}
    G &= \begin{bmatrix}
    I_{3\times 3} & I_{3\times 3} & ... & I_{3\times 3}\\
    \lfloor x_1 \rfloor_\times & \lfloor x_2 \rfloor_\times & ... & \lfloor x_n \rfloor_\times
    \end{bmatrix}, \label{eq_G}\\
    \lfloor x_i \rfloor_\times &= \begin{bmatrix}
    0 & -x_i^{(3)} & x_i^{(2)}\\
    x_i^{(3)} & 0 & -x_i^{(1)}\\
    -x_i^{(2)} & x_i^{(1)} & 0
    \end{bmatrix}.
\end{align}

Here, the form of \(\lfloor x_i \rfloor_\times\) ensures \( \lfloor x_i \rfloor_\times f_i = x_i \times f_i \). In \cref{eq_c1}, \(\epsilon\) is a small constant and \(A \succeq B\) indicates \(A-B\) is positive semi-definite. \cref{eq_c1} asserts that \(G\) is full-rank; \cref{eq_c2} asserts that contact forces balance each other; \cref{eq_c3} asserts that \(f_i\) stays within its friction cone; and \cref{eq_c4} asserts that contact points lie on the object surface.

To satisfy these constraints, one must solve for \(\{f_i\}\) that meet \cref{eq_c2,eq_c3}, a time-consuming process. To expedite this, we simplify these equations into:
\begin{subequations}
    \begin{align}
        Gf = G(f^n+f^t) = 0,\\
        G\frac{f^n}{\Vert f^n\Vert _2} = -\frac{Gf^t}{\Vert f^n\Vert _2},\\
        Gc = -\frac{Gf^t}{\Vert f^n\Vert _2},
    \end{align}
    \label{eq_3c}%
\end{subequations}%
where \(f^n\) and \(f^t\) represent the normal and tangential components of \(f\), and \(c = [c_1^T c_2^T \ldots c_n^T]^T\) represents friction cone axes. We approximate \(Gf\) with \(Gc\), which consists of object surface normals at each \(x_i\). This simplifies \cref{eq_c} into:
\begin{subequations}
    \begin{align}
        GG' &\succeq \epsilon I_{6\times 6}, \label{eq_2c1}\\
        \Vert Gc\Vert_2 & < \delta, \label{eq_2c2}\\
        x_i &\in S(O), \label{eq_2c3}
    \end{align}
    \label{eq_2c}%
\end{subequations}%
where \(\delta\) is the maximum allowed error due to our relaxation. Using \cref{eq_2c}, solving for \(f\) becomes unnecessary. The constraints for \(x_i\) turn quadratic, significantly accelerating force-closure verification. The residual \(\Vert Gc\Vert_2\) accounts for discrepancies between contact forces and friction cone axes.

We further rewrite \cref{eq_2c} as soft constraints for gradient-based optimization:
\begin{equation}
    FC(x,O) = \lambda_0(GG' - \epsilon I_{6\times6}) + \Vert Gc\Vert _2 + w\sum_{x_i\in x} \max(d^\mathrm{SDF}_O(x_i), 0),
    \label{eq_fc_soft}
\end{equation}
where $\lambda_0(\cdot)$ gives the smallest eigenvalue, and $d^\mathrm{SDF}_O(x)$ is the \ac{sdf} from point $x$ to the object $O$, consistent with its definition in \cref{eq:energy_phy_plaus}. By minimizing $FC(x,O)$, we can find the contact points $x=\{x_i\}$ that make contact with the object while providing force-closure support on it, satisfying the constraints in \cref{eq_2c}.

\clearpage
\section{Grasp classification}\label{sec:supp:adopted-taxonomy}

To showcase the diversity of grasps generated by our system, we synthesize a total of 3,450 grasps, comprising 150 grasps with 2-5 contact points for each of the 23 objects in our study. These grasps are then categorized into 19 distinct types based on the grasp taxonomy proposed by Feix \etal~\cite{feix2015grasp}. Due to the motion limitations of \HandName{}, we make some simplifications and merge similar grasp types that share the same opposition type and thumb position. For example, we combine \textit{Prismatic 2-Finger}, \textit{Prismatic 3-Finger}, and \textit{Prismatic 4-Finger} into a single category, as they differ only in the number of fingers involved. The resulting 19 types are detailed in \ref{tab:19-types}.

\begin{figure}[t!]
    \centering
    \includegraphics[width=\linewidth]{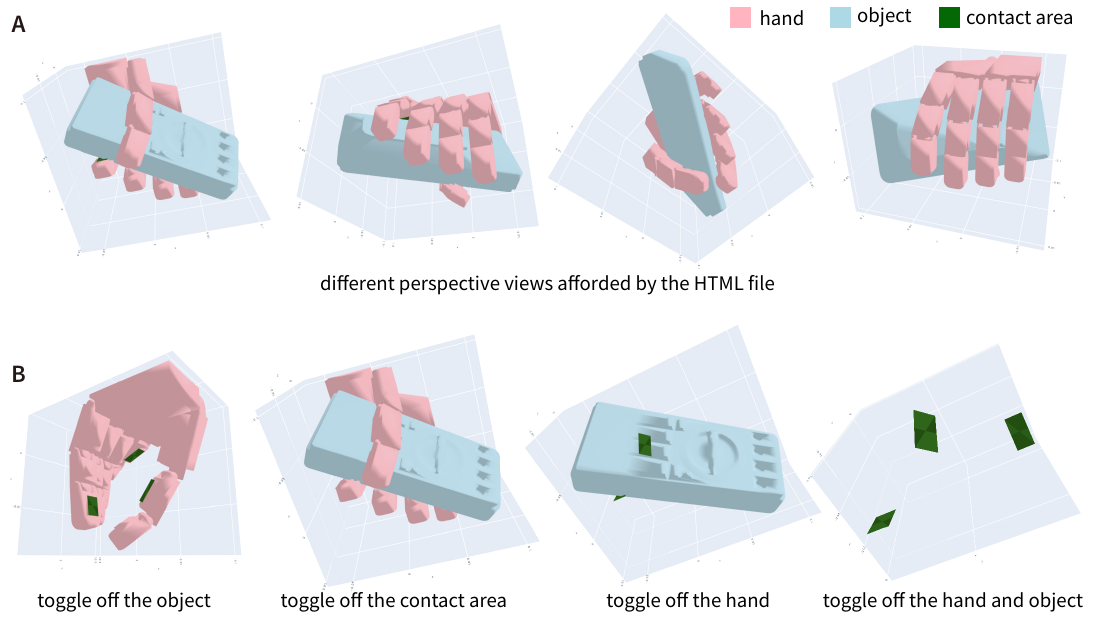}
    \caption{\textbf{\textbar~Flexible interaction for high-quality annotation:} Screenshots from an HTML file displaying all grasp-related details. The \HandName{} is shown in pink, the object (multimeter) in blue, and the contact areas in green. \textbf{(A)} Annotators can rotate the view for different perspective s. \textbf{(B)} Visibility of grasp elements can be toggled for clearer visualization. The annotation for this example is \textit{Extension Type}.}
    \label{fig:annotation-grasp-html}
\end{figure}

To ensure high-quality annotations, we enlist human annotators to label each grasp. Instead of providing a single image, we present them with an interactive HTML file that contains various grasp-related details, such as the \HandName{} configuration, object mesh, and contact areas. This interactive approach allows annotators to view the grasp from multiple perspectives and adjust the visibility of different elements, providing a more comprehensive understanding. Examples of this interactive process are visualized in \ref{fig:annotation-grasp-html}.

Given the subtle differences among the 19 grasp types, we adopt a two-step annotation process. Initially, annotators are asked to categorize the grasp into one of the three broad types (\textit{Power}, \textit{Precision}, \textit{Intermediate}) based on the definitions in Feix's taxonomy~\cite{feix2015grasp}. They then proceed to identify the specific type from the 19 available options, considering factors such as the number of contact areas and the visual similarity to examples provided in Feix \etal~\cite{feix2015grasp}.

\clearpage
\section{2D cluster map for generated grasps}\label{sec:supp:tsne-details}

The 2D map illustrated in \ref{fig:extended_algorithm} shows the human-like characteristics of the grasp synthesis algorithm. To plot this map, we adopt contact map~\cite{brahmbhatt2019contactdb,li2023gendexgrasp} and compute it over the hand surface to represent hand-object contact across different object geometries. The hand contact map \(\Omega\), defined over the hand surface \(\mathcal{S}(H)\), is computed as the distance from the hand \(H\) to the object \(O\):
\begin{equation}
    \Omega = \log (\epsilon_1 + \min(\mathbf{D} (H, O), \epsilon_2)), \label{eq:contact-map}
\end{equation}
where
\begin{equation}
    \mathbf{D} (x_h, O) = \min_{x_o \in \mathcal{S}(O)} \Vert x_h - x_o \Vert_2.
\end{equation}
The distance function \(\mathbf{D}\) measures the Euclidean distance (in meters) from any point on the hand surface \(x_h \in \mathcal{S}(H)\) to the object surface \(\mathcal{S}(O)\). To improve sensitivity in regions of close contact for better classification, we apply the \(\log\) function to curve the distance. A small \(\epsilon_1 > 0\) is used to maintain the appropriate value range for the \(\log \) function. As grasping only concerns nearby areas of the object, we introduce \(\epsilon_2 > 0\) to truncate the distance, mitigating the influence of distant areas not in contact.

In practice, the contact map is approximated with a dense point cloud with 2,170 points enveloping the hand surface. The parameters $\epsilon_1 = 0.0001, \epsilon_2 = 0.05$ are empirically chosen. Given its high dimensionality, this map is subsequently condensed to 6-dimensional space with \ac{pca} after normalization, and visualized on a 2D surface with \ac{t-SNE}. For clarity, \ref{fig:extended_algorithm} only consists of 1000 grasps, sampled using \ac{fpsampling} on the Euclidean distance of the contact maps, from all the generated grasps. The decision boundary is also determined by a \ac{svc} with the \ac{rbf} kernel using the sample points for \textit{Power} grasps and \textit{Precision grasps}, achieving an $81.54~\%$ classification accuracy.

\clearpage
\section{Quantitative analysis of grasp generation algorithm}\label{sec:supp:algorithm_detailed_results}

\begin{figure}[t!]
    \centering
    \includegraphics[width=\linewidth]{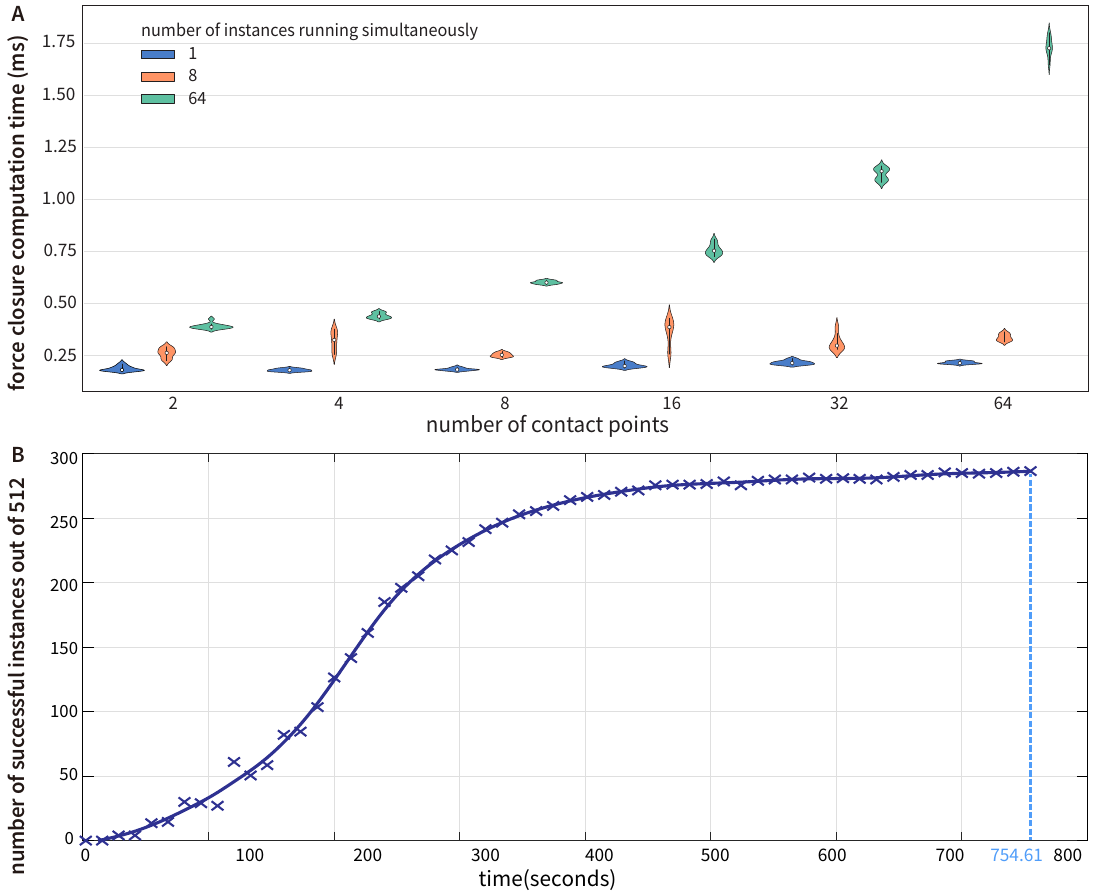}
    \caption{\textbf{\textbar~Efficient grasp repertoire generation.} \textbf{(A)} Our method rapidly estimates force closure errors using \cref{eq_fc_soft}. Even with 64 contact points across 64 parallel instances, each test takes under $2~\mathrm{ms}$. \textbf{(B)} Utilizing \ac{mala}, the algorithm navigates the complex energy landscape efficiently. In a test with 512 parallel instances for grasping a ball, the entire process takes just $754.61~\mathrm{s}$, yielding 287 successful grasps at a $56~\%$ success rate.}
    \label{fig:quantitative_result_of_algorithm}
\end{figure}

Our proposed algorithm efficiently estimates force closure errors using \cref{eq_fc_soft}, addressing the most time-consuming aspect of traditional grasp generation methods~\cite{sahbani2012overview,rodriguez2012caging,prattichizzo2012manipulability,prattichizzo2016grasping,murray2017mathematical}. We validate its efficiency by running experiments on an Intel Xeon CPU ($2.90~\mathrm{GHz}$) coupled with a single NVIDIA RTX 3090 GPU. Various combinations of contact points and parallel instances are tested, and the results are shown in \ref{fig:quantitative_result_of_algorithm}A. Notably, even the most complex scenario involving 64 contact points across 64 parallel instances takes under $2~\mathrm{ms}$ per test.

For gradient-based methods, navigating a complex, non-convex energy landscape without getting stuck in local minima is a significant challenge. Our method addresses this issue effectively through \ac{mala}. We demonstrate this by running 512 grasp generation instances for grasping a ball on the same hardware setup. The results, presented in \ref{fig:quantitative_result_of_algorithm}B, reveal that the entire process takes just $754.61~\mathrm{s}$ and yields 287 successful grasps, achieving a $56~\%$ success rate. This efficiency in generating a diverse set of grasps is a direct result of our approach.

\renewcommand{\figurename}{Fig.~}
\clearpage

\clearpage

\clearpage

\clearpage

\clearpage
\begin{sidewaystable}[t!]
    \centering
    \caption{Quantitative comparison of \HandName{} with state-of-the-art tactile sensing arrays across multiple performance dimensions.}
    \label{tab:comparison-eskin}
    \begin{tabular}{lcccccccc}
        \toprule
        \multirow{2}{*}{\textbf{System}} & \multirow{2}{*}{\shortstack{\textbf{Sensing}\\\textbf{area}}} & \multirow{2}{*}{\textbf{Resolution}$\uparrow$} & \multirow{2}{*}{\shortstack{\textbf{Transduction}\\\textbf{method}}} &\multirow{2}{*}{\shortstack{\textbf{Wiring}\\\textbf{complexity}}} & \multirow{2}{*}{\shortstack{\textbf{Cost}\\\textbf{per sensor}}} \multirow{2}{*}{$\downarrow$} &\multirow{2}{*}{\shortstack{\textbf{Sensitivity}\\\(\boldsymbol{(\mathrm{Kpa}^{-1})}\)}}\\\\
        \midrule
        Yu et al.~\cite{yu2022all} & \textbf{Full hand} & 21 taxels/hand & Piezoresistive & Per taxel & --- & ---\\
        Luo et al.~\cite{luo2021learning} & \textbf{Full hand} & \num{4.5} taxels/\(\mathrm{cm}^2\) & Piezoresistive & Per taxel & --- & ---\\
        Li et al.~\cite{li2020skin} & Fingers & \num{10} taxels/hand & Piezothermic & Per taxel & --- &$117~\mathrm{mV}$\\
        Lee et al.~\cite{lee2019neuro} & \textbf{Full hand} & \num{23} taxels/\(\mathrm{cm}^2\) & Resistive & Per taxel & --- & ---\\
        Sundaram et al.~\cite{sundaram2019learning} & \textbf{Full hand} & \num{548} taxels/hand & Piezoresistive & Per taxel & $\$110$ & ---\\
        Kim et al.~\cite{kim2014stretchable} & \textbf{Full hand} & \num{2} taxels/\(\mathrm{cm}^2\) & Piezoresistive & Per taxel & --- & $0.41$\\
        \textbf{\HandName{}} & \textbf{Full hand} & \(\boldsymbol{10~000}\) \textbf{taxels/}\(\boldsymbol{\mathrm{cm}^2}\) & Vision-based & \textbf{Per sensor} & \(\boldsymbol{\$20}\) & $0.48$\\
        \bottomrule
    \end{tabular}
\end{sidewaystable}

\clearpage
\begin{table}[t!]
    \centering
    \small
    \caption{\textbf{Dimensional specifications of the six tactile sensor configurations implemented across different functional regions of the hand.}}
    \label{tab:extended_dimension}
    \begin{tabular}{lcccccc}
        \toprule
        \textbf{Sensor configuration} &\textcircled{a} &\textcircled{b} &\textcircled{c} &\textcircled{d} &\textcircled{e} & \textcircled{f}\\
        \midrule
        \textbf{Number of camera} &\(1\) &\(1\) &\(1\) &\(1\) &\(1\) &\(2\)\\
        \multirow{2}{*}{\shortstack{\textbf{Physical dimension}\\(\(h~\textrm{mm}\times w~\textrm{mm}\))}} & \multirow{2}{*}{\(33\times20\)} & \multirow{2}{*}{\(29\times20\)} & \multirow{2}{*}{\(41\times20\)} & \multirow{2}{*}{\(45\times25\)} & \multirow{2}{*}{\(48\times29\)} & \multirow{2}{*}{\(69\times47\)} \\
        \\
        \bottomrule
    \end{tabular}
\end{table}

\clearpage

\begin{table}[t!]
    \centering
    \small
    \caption{Quantitative similarity comparison between the real and the simulated sensor outputs.}
    \label{tab:quant_gelsim}
    \begin{tabular}{ccc}
        \toprule
        \textbf{L1} & \textbf{SSIM} & \textbf{PSNR} \\
        \midrule
        $0.009\pm0.003$ & $0.976\pm0.008$ & $38.808\pm1.236$\\
        \bottomrule
    \end{tabular}
\end{table}

\clearpage

\begin{table}[t!]
    \centering
    \footnotesize
    \caption{The 19 distinct grasping poses utilized in our classification scheme.}
    \label{tab:19-types}
    \begin{tabular}{c c c}
    \toprule
    \textbf{Grasp Type} & \textbf{Included Types in~\cite{feix2015grasp}} & \textbf{Type} \\
    \midrule
    \textit{Adducted Thumb} & \textit{Adducted Thumb}, \textit{Light Tool} & Power \\
    \textit{Fixed Hook} & \textit{Fixed Hook} & Power \\
    \textit{Index Finger Extension} & \textit{Index Finger Extension} & Power \\
    \textit{Ring} & \textit{Ring} & Power \\
    \textit{Power Disk} & \textit{Power Disk} & Power \\
    \textit{Extension Type} & \textit{Extension Type} & Power \\
    \textit{Palmar} & \textit{Palmar} & Power \\
    \textit{Heavy Wrap} & \textit{Large Diameter}, \textit{Small Diameter}, \textit{Medium Wrap} & Power \\
    \textit{Power Sphere} & \textit{Sphere 3-Finger}, \textit{Sphere 4-Finger}, \textit{Power Sphere} & Power \\
    \textit{Distal Type} & \textit{Distal Type} & Power \\
    \textit{Stick} & \textit{Stick}, \textit{Ventral} & Intermediate \\
    \textit{Lateral} & \textit{Lateral} & Intermediate \\
    \textit{Lateral Tripod} & \textit{Lateral Tripod} & Intermediate \\
    \textit{Prismatic} & \textit{Prismatic 2-Finger}, \textit{Prismatic 3-Finger}, \textit{Prismatic 4-Finger} & Precision \\
    \textit{Writing Tripod} & \textit{Writing Tripod} & Precision \\
    \textit{Precision Disk} & \textit{Precision Disk} & Precision \\
    \textit{Pincer} & \textit{Palmar Pinch}, \textit{Tip Pinch}, \textit{Inferior Pincer} & Precision \\
    \textit{Precision Sphere} & \textit{Tripod}, \textit{Quadpod}, \textit{Precision Sphere} & Precision \\
    \textit{Parallel Extension} & \textit{Parallel Extension} & Precision \\
    \bottomrule
    \end{tabular}
\end{table}

\end{document}

%% file: config.tex
\acrodef{gep}[GEP]{Generalized Earley Parser}
\acrodef{t-aog}[T-AOG]{temporal And-Or Graph}
\acrodef{dps}[DPS]{Deep Photometric Stereo}
\acrodef{mala}[MALA]{Metropolis-Adjusted Langevin Algorithm}
\acrodef{mcp}[MCP]{metacarpophalangeal}
\acrodef{pip}[PIP]{proximal interphalangeal}
\acrodef{dip}[DIP]{distal interphalangeal}
\acrodef{adelm}[ADELM]{Attraction-Diffusion Energy Landscape Mapping}
\acrodef{dof}[DoF]{Degree-of-Freedom}
\acrodefplural{dof}[DoFs]{Degrees of Freedom}
\acrodef{mep}[MEP]{Minimum Energy Path}
\acrodef{pca}[PCA]{Principle Component Analysis}
\acrodef{svc}[SVC]{Support Vector Classifier}
\acrodef{fpsampling}[FPS]{Furthest Point Sampling}
\acrodef{t-SNE}[t-SNE]{t-distributed Stochastic Neighbour Embedding}
\acrodef{rbf}[RBF]{Radial Basis Function}
\acrodef{f-th}[F-TOUCH Hand]{anthropomorphic hand with Full-covered TOUCH sensing}
\acrodef{ai}[AI]{artificial intelligence}
\acrodef{sdf}[SDF]{Signed Distance Function}
\acrodef{ffc}[FFC]{flexible flat cable}
\acrodef{llm}[LLM]{large language model}
\newcommand{\HandName}{F-TAC Hand}
\newcommand{\suppUrlnmi}{https://vimeo.com/1039184307}

\newcommand{\Title}{Embedding high-resolution touch across robotic hands enables adaptive human-like grasping}
\newcommand{\TitleSupp}{Embedding high-resolution touch across robotic hands enables adaptive human-like grasping}

\hypersetup{
  colorlinks=true,
  urlcolor=blue
}